\pdfoutput=1
\documentclass[
  10pt,
  logo,
  onecolumn,
  copyright,
  nodate
]{nvidiatechreport}

\usepackage[numbers,sort,square]{natbib}
\usepackage{flafter}
\usepackage{float}
\usepackage{multirow}

\renewcommand*{\backref}[1]{}
\renewcommand*{\backrefalt}[4]{}
\renewcommand{\headingfont}{\color{black}\bfseries\fontsize{12}{13}\selectfont}

\titlespacing*{\section}{0pc}{2.11ex plus 3.1pt minus 2.3pt}{5pt}
\titlespacing*{\subsection}{0pc}{1.77ex plus 2.3pt minus 1.5pt}{2pt}
\makeatletter
\newskip\figureafterskip
\def\@figurecaptype{figure}
\renewcommand\float@endH{%
  \@endfloatbox\vskip\intextsep
  \if@flstyle\setbox\@currbox\float@makebox\columnwidth\fi
  \box\@currbox
  \ifx\@captype\@figurecaptype
    \vskip\figureafterskip
  \else
    \vskip\intextsep
  \fi
  \relax}
\makeatother

\hypersetup{
  pdftitle={Reasoning Supervision and Capability Retention in LoRA Domain Adaptation},
  pdfauthor={Saleh Ashkboos, Dhruv Desai, Lavinia Ghita}
}

\title{Reasoning Supervision and Capability Retention
in LoRA Domain Adaptation}
\author{\large Saleh Ashkboos, Dhruv Desai, Lavinia Ghita\\
\normalfont NVIDIA}
\date{}

\begin{document}

\maketitle

\noindent\textbf{Abstract.}\enspace
Fine-tuning foundation models on financial data has a wide range of applications, including identifying event types from news, monitoring risk and compliance language, and answering questions over filings and earnings calls.
In this setting, the main challenge is to improve task performance without degrading the
generalization capabilities of the model. In this paper, we present an extensive set of
ablations for fine-tuning Qwen3-8B with low-rank adaptation (LoRA). We compare three training regimes:
(1) \emph{response-only} LoRA, where the supervised response is the label alone;
(2) \emph{single-adapter reasoning}, where reasoning traces from a teacher
model are included alongside the labels in the training loss; and
(3) \emph{dual-adapter composite} training, where separate LoRA adapters are
trained on traces and labels.
Response-only LoRA improves in-domain performance and leaves general capabilities near those of the frozen model. The in-domain change is largest for calibration and argmax accuracy. Under single-adapter reasoning, the in-domain metrics remain behind those of the response-only regime, while MMLU-Pro can rise slightly and the other retention benchmarks stay within ordinary variation. Dual-adapter composite training remains on the same side of this comparison and yields no further gain.
Finally, we compress the teacher traces by summarizing them before training,
and show that shorter summaries move calibration toward the response-only regime without reaching it.

\section{Introduction}\label{sec:introduction}

Low-rank adaptation (LoRA)~\citep{lora} is a widely used fine-tuning method for
resource-constrained settings. For a weight matrix $\mathbf{W} \in \mathbb{R}^{m \times n}$
of a linear layer, LoRA keeps $\mathbf{W}$ frozen and parameterizes its update as the scaled
low-rank product $\Delta\mathbf{W} = (\alpha/r)\,\mathbf{A}\mathbf{B}$, where
$\mathbf{A} \in \mathbb{R}^{m \times r}$ and $\mathbf{B} \in \mathbb{R}^{r \times n}$ are the
trainable factors (or adapters), $r \ll \min(m, n)$ is the rank, and $\alpha$ is a scaling coefficient.
This reduces the number of trainable parameters per layer from $mn$ to $r(m+n)$, which is
what makes fine-tuning feasible on limited hardware.

LoRA makes domain specialization computationally cheap. Yet specialization is only
useful if the adapted model keeps its broad capabilities. After fine-tuning, it should
still generalize to unseen problems, follow instructions, and express calibrated
uncertainty. Retaining these capabilities is especially challenging for classification-like tasks
with long inputs. In this setting, the adapter must learn a specialized decision
boundary while preserving the base model's general reasoning.

One potential solution to this problem is to distill knowledge from a larger, more capable model
(the \emph{teacher}) into the \emph{student} during fine-tuning. Teacher-generated
reasoning traces show the intermediate steps that lead to the final answer and thus
provide far more supervision per example than the answer alone. However, these traces
also contain repetition, hedging, self-correction, and prose specific to the teacher,
and training on full traces asks the student to reproduce all of it. Reasoning traces
therefore offer a rich domain signal, but whether the student keeps its general
capabilities depends on how this signal changes the model during fine-tuning.

Training on traces from a larger model is slow, because each example contains many
more tokens than the answer alone. This makes trace-based fine-tuning hard to run in
low-resource settings, which are the main use case for LoRA. One way to address this
is to summarize the teacher traces before they are used to fine-tune the student,
so that the student is trained on only a fraction of the original tokens. However,
the choice of compression method and compression ratio is critical: too much
compression can remove the information that makes the traces useful and leave
fine-tuning with little to learn from.

In this study, we examine how supervising with the response (or label) alone, compared with
supervising with a teacher reasoning trace, affects domain performance and capability
retention. We study two ways of incorporating traces into LoRA fine-tuning. The first is a
blended loss, where a single adapter is trained on a weighted combination of the trace loss
and the label loss, with a weight that controls how much each part contributes. The second
is dedicated adapters, where the rank budget is split between one adapter for the trace and
one for the label, and the two are merged into a single update at inference. The trace is
either the full teacher trace or a summary of it, and we vary the degree of compression.
All comparisons use matched training configurations, so that a consistent trend can be
separated from a single favorable setting.

\paragraph{Contributions.} Our main contributions are as follows:
\begin{enumerate}[topsep=-\parskip,partopsep=0pt]
  \item We present a systematic study of reasoning supervision in LoRA fine-tuning
        for a financial classification task. We use a dataset of 200K examples as our main training set.

  \item To this end, we first build a strong response-only baseline by tuning the LoRA hyperparameters of a
        Qwen3-8B \citep{qwen3} model on our dataset.

  \item We study how to incorporate the reasoning traces of a larger teacher model into
        LoRA fine-tuning through a blended label/trace objective, and we ablate the
        weight that balances the two terms.

  \item We propose a composite adapter that uses separate LoRA modules for the labels and
        the traces and ablate the allocation of ranks between the two adapters under a
        fixed rank budget.

  \item We summarize the teacher traces before training and ablate the effect of several
        compression levels.

  \item We evaluate all configurations on both the in-domain task and a set of retention
        benchmarks, so that gains on the target task can be weighed against losses in
        general capabilities.
\end{enumerate}

\vspace{\baselineskip}
The rest of the paper is organized as follows.
Section~\ref{sec:related} reviews related work.
Section~\ref{sec:methods} describes the dataset, the supervision methods, and the staged protocol.
Section~\ref{sec:experiments} reports the results.
Section~\ref{sec:conclusion} presents the conclusions.
Appendix~\ref{app:appendix} contains the hyperparameter search (Appendix~\ref{app:hparam_search}),
the detailed results for the response-only, single-adapter reasoning, and dual-adapter composite stages (Appendices~\ref{app:response_only_results},
\ref{app:single_adapter_results}, and~\ref{app:dual_adapter_results}), the trace-compression prompt and the resulting summary lengths
(Appendix~\ref{app:compression_details}), and the detailed results for the summarized-traces stage (Appendix~\ref{app:compression_results}).

\section{Related Work}\label{sec:related}

Low-rank adaptation (LoRA) updates a frozen model through trainable low-rank
matrices~\citep{lora}, while QLoRA extends this setup to quantized base
models~\citep{dettmers2023qlora}. A small trainable footprint does not, however,
guarantee retention. Compared with full fine-tuning, LoRA often learns less and
forgets less, with rank controlling the tradeoff~\citep{biderman2024lora};
forgetting also grows with the number of updated parameters and optimization
steps~\citep{kalajdzievski2024forgetting}. With sufficient capacity, broad layer
coverage, and appropriate tuning, LoRA can nevertheless match full
fine-tuning~\citep{schulman2025lora}. This dependence on the training regime
motivates evaluating adaptation and retention jointly rather than treating LoRA
as a capability-preserving baseline.

Prior financial language models span mixed financial and general pretraining in
BloombergGPT~\citep{wu2023bloomberggpt}, efficient specialization in
FinGPT~\citep{yang2023fingpt}, and instruction and evaluation resources such as
FinMA and FLARE~\citep{xie2023pixiu}. The closest work to our setting studies
label-only and blended chain-of-thought supervision for LoRA and logit
distillation on pruned students~\citep{ghita2026scaling}. At a fixed student
size, blended LoRA is stable across three seeds and surpasses blended
chain-of-thought distillation on MMLU and MMLU-Pro at the largest data budgets.
Its classification gains plateau, while gold-label NLL and Brier score first
improve and then worsen. We instead keep a dense Qwen3-8B backbone fixed and
compare response-only LoRA, in which the response is the label alone;
single-adapter reasoning; and dual-adapter composite LoRA under matched budgets.

Our reasoning supervision can also be understood as sequence distillation.
Classical knowledge distillation matches softened teacher and student
distributions~\citep{hinton2015distilling}; sequence distillation replaces that
distribution with the teacher's highest-scoring decoded
response~\citep{kim2016sequence}. Our sampled traces are likewise hard sequence
targets. Prior rationale distillation has trained rationale generation and label
prediction as separate tasks~\citep{hsieh2023distilling}, whereas we train on a
single response containing a trace followed by its label. We treat the trace as
a verbalized rationale rather than as a faithful account of the teacher's
internal process.

Which rationales help depends on both the student and the objective. The useful
level of granularity varies with student capacity, and stronger teachers do not
always yield better students~\citep{chen2025keyfactors}. Existing methods have
separated rationale and answer learning~\citep{dai2024cascod} or combined LoRA
experts trained on different reasoning tasks~\citep{li2024modecotd}. Our
narrower intervention trains two always-active additive factors on a single
response containing a trace and a label. The trace factor receives both losses,
the label factor receives only the label loss, and the total rank is matched to
the single-adapter control. Because the factors merge into one static update,
the intervention changes the training procedure rather than the inference
architecture.

Trace length introduces a related efficiency tradeoff. Prior work shows that
full traces give the highest accuracy, while compressed traces reduce training
and inference cost and usually outperform truncation at the same
length~\citep{griot2026compress}. We therefore compare full traces with semantic
summaries at several compression levels, rather than assuming that shorter
traces are inherently better or worse.

\begin{figure}[!t]
  \centering
  \includegraphics[width=\linewidth]{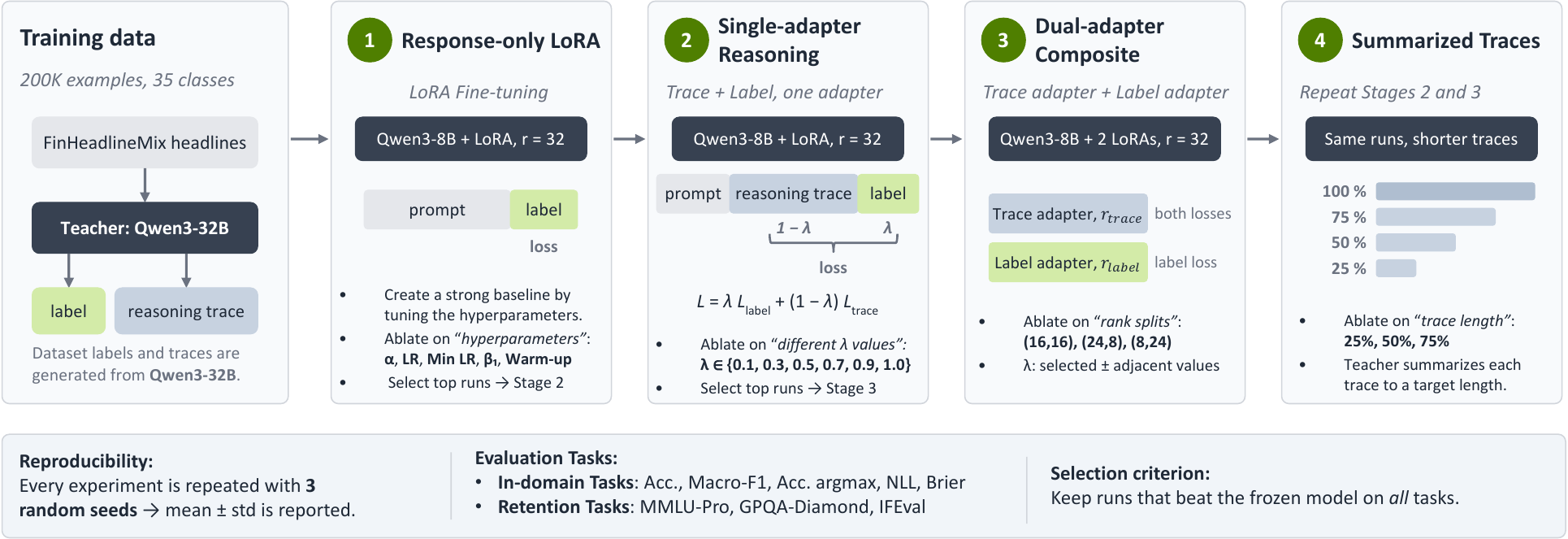}
  \caption{Overview of the staged protocol.
  The teacher (Qwen3-32B) labels 200K FinHeadlineMix headlines and generates a reasoning trace for each.
  We then fine-tune the student (Qwen3-8B) with LoRA in four stages:
  (1) response-only LoRA, trained on the label alone over a hyperparameter grid,
  (2) single-adapter reasoning, trained on the trace and the label with the blended loss of Eq.~\eqref{eq:region_loss},
  (3) dual-adapter composite LoRA, which splits the rank budget between a trace adapter and a label adapter,
  and (4) summarized traces, which repeats Stages~2 and~3 with teacher-generated summaries at target lengths of 25\%, 50\%, and 75\%.
  In every stage, each run uses three seeds and is evaluated on the in-domain task and the retention benchmarks.
  A run is carried to the next stage when it outperforms the frozen model on every task. If none does, we fall back to the run with the highest average retention score among those that beat the frozen model on the in-domain tasks.}
  \label{fig:protocol_overview}
\end{figure}

\section{Methods of LoRA Supervision}\label{sec:methods}

In this section, we first describe the models and dataset used in our study, along with
practical considerations such as hyperparameter tuning. We then describe the supervision
methods used in LoRA fine-tuning. Finally, we present the staged protocol that organizes
our study.

\subsection{Dataset Generation and Model Architecture}\label{sec:method_dataset_model}

We study LoRA fine-tuning in the financial domain. To this end, we fine-tune Qwen3-8B~\citep{qwen3}
on a training set of 200K examples whose inputs are headlines from the FinHeadlineMix
dataset~\citep{finheadlinemix}.\footnote{\url{https://huggingface.co/datasets/nvidia/FinHeadlineMix}}
As supervision, we use the labels and reasoning traces generated by Qwen3-32B for the same inputs.
In the rest of the paper, we refer to the 8B and 32B models as the ``student'' and
the ``teacher'', respectively. The choice of teacher and dataset size follows~\citep{ghita2026scaling}.

\subsection{Response-only LoRA}\label{sec:method_response_only}

In the first stage, we apply LoRA fine-tuning to the student using only the labels from our dataset. This
allows us to assess how training on the labels affects domain performance and retention on MMLU-Pro, GPQA, and IFEval.
In this stage, we also perform hyperparameter tuning to find the best
settings, which we then use in the subsequent stages. This establishes how much domain utility and
calibration the available labels can provide while preserving the model's held-out capabilities.
Across all experiments, the LoRA rank is \(r = 32\).

\subsection{Single-adapter Reasoning}\label{sec:method_single_adapter}

In the second stage, we apply a reasoning method which keeps a single adapter and changes only the supervision signal.
For traced examples, the teacher response is serialized
as a reasoning region followed by the final label. The adapter is trained with a per-example,
region-balanced loss
\begin{equation}
  \mathcal{L}_{i}
  = \lambda\,\overline{\mathcal{L}}_{\mathrm{label},i}
  + (1-\lambda)\,\overline{\mathcal{L}}_{\mathrm{trace},i},
  \label{eq:region_loss}
\end{equation}
where each bar denotes the mean cross-entropy over the tokens in that region and \(\lambda\) controls the weight
given to the label. Because the loss is averaged within each region, the weight of the trace does not depend on its length:
a long chain of thought cannot dominate the label simply by having more tokens.
A high \(\lambda\) focuses training on the label, while a low \(\lambda\) encourages imitating the
teacher's reasoning. In our experiments, we ablate a range of \(\lambda\) values to study the role of
traces during fine-tuning. We note that \(\lambda=1\) is equivalent to response-only training in terms of
the loss in Eq.~\eqref{eq:region_loss}. However, the two settings are not identical: although the losses
are equivalent, both the trace and the label tokens are passed to the model when \(\lambda=1\) (see
Section~\ref{sec:experiments} for details).

\subsection{Dual-adapter Composite LoRA}\label{sec:method_dual_adapter}

The third stage applies a second reasoning method, which uses the same serialized regions and the same blended loss
(Section~\ref{sec:method_single_adapter}) but splits the rank budget across two adapters within a single model: a
trace adapter and a label adapter. Both adapters are active on every token and are merged into one static update at inference. The trace adapter is trained by the full blended loss, and the label adapter is trained by the label term only. This comparison keeps the total rank, training budget, and evaluation
protocol fixed relative to single-adapter reasoning.
For a given rank budget \(r\), we evaluate three cases:

\begin{itemize}
  \item \textbf{Equal Split}, \((r_{\mathrm{trace}},r_{\mathrm{label}})=(r/2,r/2)\): the rank is
  divided evenly between the trace and label adapters;
  \item \textbf{Trace-Heavy Split}, \((r_{\mathrm{trace}},r_{\mathrm{label}})=(3r/4,r/4)\): more rank
  is allocated to the trace adapter;
  \item \textbf{Label-Heavy Split}, \((r_{\mathrm{trace}},r_{\mathrm{label}})=(r/4,3r/4)\): more rank
  is allocated to the label adapter.
\end{itemize}

In all cases, the ratio \(\alpha/r\) is held fixed for each adapter.
For a given value of \(\lambda\), selected from the top-performing runs of the previous stage, we
ablate all three configurations to evaluate the rank allocation in the dual-adapter setup.

\subsection{Summarized Traces}\label{sec:method_summarized_traces}

In the fourth stage, we repeat our reasoning experiments with summarized traces, for both the single and the composite adapter.
Following~\citep{griot2026compress}, we summarize each trace with our 32B teacher model to a target token budget. To this end, each trace
is wrapped in a prompt that asks the model to rewrite it as a summarized version of a given length.
We ablate several length ratios to measure the effect of summarization on the in-domain and retention performance of the fine-tuned model.
In each case, we allow the model to generate up to $10\%$ (with a constant offset) more tokens than the target,
so that the summary is not cut off before it is complete.
The implementation details and the results of this ablation are presented in Section~\ref{sec:exp_summarized_traces}.

\subsection{Sequential Experimental Design}\label{sec:method_protocol}

Comparing fine-tuning schemes is only meaningful if differences in accuracy reflect the schemes themselves rather than the hyperparameters chosen for any one of them.
However, a full hyperparameter search is impractical because the search space grows exponentially.
We therefore organize the study in stages, each building on the runs selected in the previous one.
We first evaluate the frozen model without fine-tuning and use it as the primary baseline.
The fine-tuning experiments then proceed in four stages:

\begin{enumerate}[label=\textbf{Stage~\arabic*.}, leftmargin=*]
  \item We run response-only LoRA (Section~\ref{sec:method_response_only}) over the selected hyperparameter grid and keep the runs that satisfy the selection criterion. Details of the hyperparameter selection are given in Appendix~\ref{app:hparam_search}.
  \item Starting from these runs, we train single-adapter reasoning LoRA (Section~\ref{sec:method_single_adapter}) over a range of $\lambda$ values in Eq.~\eqref{eq:region_loss} and keep the top runs, satisfying the selection criterion.
  \item For each run kept in Stage~2, we train dual-adapter composite LoRA (Section~\ref{sec:method_dual_adapter}) with equal, trace-heavy, and label-heavy splits. We repeat these experiments at the $\lambda$ values adjacent to the selected one, so that the final comparison does not depend on the exact $\lambda$ chosen in Stage~2.
  \item Finally, we repeat the experiments of Stage~2 and Stage~3 with summarized traces, where each trace in the training data is replaced by its summary (Section~\ref{sec:method_summarized_traces}).
  We ablate different summarization levels in this stage.
\end{enumerate}
\vspace{-\topsep-\partopsep-\parskip}%
Figure~\ref{fig:protocol_overview} summarizes the above stages in our study.

\paragraph{Selection criterion.}
At every stage we apply the same rule: a run is kept only if it outperforms the frozen baseline on \emph{all} tasks, both in-domain tasks and retention (general-capability) tasks.
We use this rule to satisfy an important practical consideration: a model that gains in-domain accuracy at the cost of its general capabilities is of limited use.
Using one rule throughout also keeps the comparison fair across schemes.
When no run satisfies this rule, we fall back to the run with the highest average retention score among those that outperform the baseline on all in-domain tasks.
Full task details are given in Section~\ref{sec:exp_setup}.

\section{Experimental Results}\label{sec:experiments}

This section presents the experimental results of our study.
For each experiment stage, we first describe the setup and methodology,
then report the results and discuss the findings.

\subsection{Training and Evaluation Setup}\label{sec:exp_setup}

All experiments fine-tune a single base model, Qwen3-8B, with LoRA on a dataset of headline inputs from FinHeadlineMix~\citep{finheadlinemix}, annotated by the teacher model (see Section~\ref{sec:method_dataset_model}).
We train every model with AdamW~\citep{adamw} for a single epoch, repeat each configuration with three random seeds, and report the mean and standard deviation across seeds. The training hyperparameters are ablated in Section~\ref{sec:exp_response_only}.
We evaluate every fine-tuned model along two axes:
(i)~\textbf{In-domain} and (ii)~\textbf{Retention}.

\paragraph{In-domain Tasks.} We measure in-domain performance on a held-out, class-stratified set of 10,000 financial-headline event-classification examples.
We report \textbf{greedy accuracy} (``Acc'') and \textbf{macro-F1} (``F1'') over the 35 event classes.
To assess calibration, we additionally report the \textbf{negative log-likelihood} (``NLL'') and \textbf{Brier score}~\citep{brier} (``Brier'')
of the model's label distribution, together with \textbf{argmax accuracy} (``Acc.\ argmax'').
Argmax accuracy scores each of the 35 class strings as a teacher-forced continuation of the prompt
and counts a prediction as correct when the gold class receives the highest likelihood.
Unlike greedy accuracy, which parses the label from the model's greedy generation and therefore also penalizes formatting failures,
argmax accuracy isolates how well the model's class-token distribution ranks the correct label (it is computed from the same distribution as NLL and Brier).

\paragraph{Retention of General Capabilities Tasks.} We measure retention on three public benchmarks: \textbf{MMLU-Pro}~\citep{mmlupro} for broad multi-task knowledge and reasoning,
\textbf{GPQA-Diamond}~\citep{gpqa} (``GPQA'') for graduate-level scientific reasoning,
and \textbf{IFEval}~\citep{ifeval} (``IFEval'') for instruction following, for which we report prompt-level strict accuracy.
All evaluations use greedy decoding with the model's thinking mode disabled and the same prompt template as in training,
so that scores are directly comparable across all fine-tuning configurations and against the untuned base model.

\subsection{Baseline Results}\label{sec:exp_baseline}

Table~\ref{tab:baseline} reports the performance of the untuned base model (Qwen3-8B without fine-tuning) on all metrics.
Each evaluation is run three times, and we report the mean and standard deviation (std).
The largest run-to-run variation is on GPQA (std $\approx$ 4.3\%) among the retention benchmarks and on macro-F1 (std $\approx$ 1.1\%) among the in-domain metrics.
GPQA contains only 198 questions, so each one is worth roughly 0.5 percentage points, and the questions are difficult enough that the model sits near the decision boundary on many of them, so the
score moves noticeably.
Macro-F1 is similarly sensitive because it weights every class equally regardless of its size: for small classes, a few flipped predictions can shift that class's F1 substantially.

\begin{table}[H]
\centering
\caption{In-domain and retention results for the frozen Qwen3-8B model over three independent evaluation runs (mean $\pm$ std). ($\uparrow$): higher is better, ($\downarrow$): lower is better.}
\label{tab:baseline}
\begin{tabular}{llccc|c}
\toprule
Task Type & Metric & Run 1 & Run 2 & Run 3 & Mean $\pm$ Std \\
\midrule
\multirow{5}{*}{In-domain}
  & NLL ($\downarrow$)         & 5.6993 & 5.7008 & 5.7033 & $5.7011 \pm 0.0020$ \\
  & Brier ($\downarrow$)       & 0.8595 & 0.8596 & 0.8597 & $0.8596 \pm 0.0001$ \\
  & Acc.\ argmax ($\uparrow$)  & 0.5264 & 0.5264 & 0.5260 & $0.5262 \pm 0.0002$ \\
  & Macro-F1 ($\uparrow$)      & 0.6506 & 0.6313 & 0.6521 & $0.6447 \pm 0.0116$ \\
  & Acc ($\uparrow$)           & 0.7739 & 0.7709 & 0.7731 & $0.7726 \pm 0.0016$ \\
\midrule
\multirow{3}{*}{Retention}
  & GPQA ($\uparrow$)          & 0.4798 & 0.5657 & 0.5354 & $0.5269 \pm 0.0435$ \\
  & IFEval ($\uparrow$)        & 0.8373 & 0.8558 & 0.8355 & $0.8429 \pm 0.0112$ \\
  & MMLU-Pro ($\uparrow$)      & 0.6223 & 0.6222 & 0.6229 & $0.6225 \pm 0.0004$ \\
\bottomrule
\end{tabular}
\end{table}

\subsection{Response-only LoRA Results}\label{sec:exp_response_only}

Response-only LoRA improves in-domain performance and leaves general capabilities near the frozen model. The in-domain change is largest for calibration and for which class the model ranks first. Greedy accuracy and macro-F1 move less.
We LoRA-fine-tune Qwen3-8B on this dataset with AdamW for a single epoch, and we ablate the LoRA scaling coefficient $\alpha$, the peak and minimum learning rate, the AdamW $\beta_1$, and the number of warm-up steps (see Appendix~\ref{app:hparam_search} for hyperparameter details).
We then select the \emph{top runs} that outperform the frozen base model on \emph{all} tasks, both in-domain and retention.
This criterion yields 5 runs for the 200K dataset, and these are the runs carried forward to Stage~2. 

\begin{figure}[H]
  \centering
  \includegraphics[width=\linewidth]{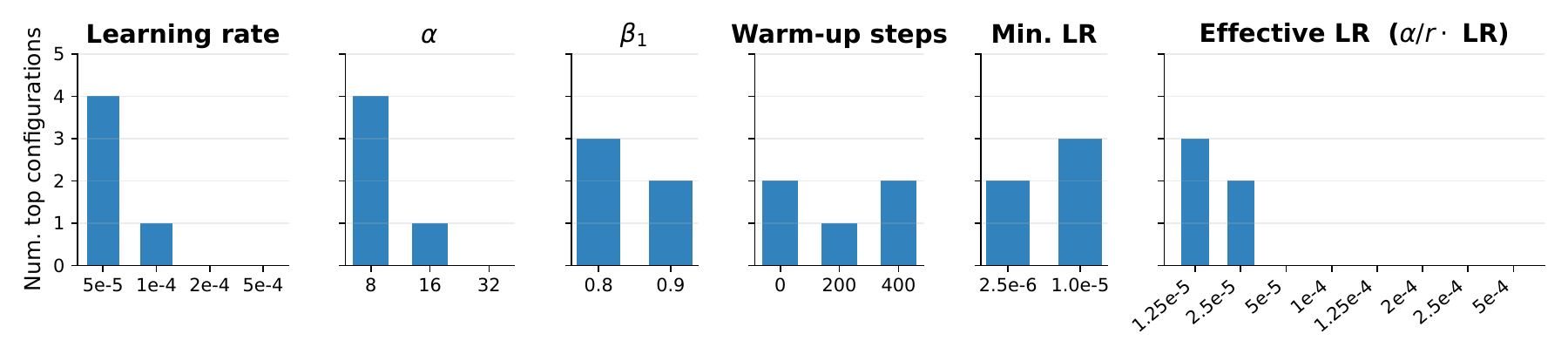}
  \caption{Distribution of the top runs over the hyperparameter grid for the 200K dataset (5 runs)
  for response-only LoRA (Stage~1).
  The last plot uses the effective step size $\alpha\eta/r$ with $r=32$, where $\eta$ is the learning rate.}
  \label{fig:hp_trends}
\end{figure}

Figure~\ref{fig:hp_trends} shows the effect of each hyperparameter on the top runs of response-only LoRA.
Learning rate and $\alpha$ have the largest impact: top runs favor low learning rates (4 of the 5 top runs use $5\times10^{-5}$) and small $\alpha$ (4 of the 5 use $\alpha = 8$, and the fifth uses $\alpha = 16$).
Consistent with this, \textit{the effective LoRA learning rate appears to be the decisive factor} as
three of the five top runs have effective step size $1.25\times10^{-5}$, and the other two sit at $2.5\times10^{-5}$.
Among the remaining parameters, $\beta_1 = 0.8$ and a minimum learning rate of $1\times10^{-5}$ give better results, while the number of warm-up steps has little effect.
We present the complete top-run results in Appendix~\ref{app:response_only_results}.

\subsection{Single-adapter Reasoning Results}\label{sec:exp_single_adapter}

Single-adapter reasoning on full traces leaves the in-domain metrics behind the response-only run that each configuration starts from. Calibration, argmax accuracy, macro-F1, and NLL remain behind those runs. MMLU-Pro can rise slightly, while GPQA and IFEval stay within ordinary variation.
We apply the region-balanced loss (see Section~\ref{sec:method_single_adapter}) to the top runs from Stage~1.
We train each run with six different $\lambda$ values $\{0.1, 0.3, 0.5, 0.7, 0.9, 1\}$ as in Eq.~\eqref{eq:region_loss}
and select the top runs that satisfy our criterion (defined in Section~\ref{sec:method_protocol}).

\begin{figure}[!t]
  \centering
  \includegraphics[width=\linewidth]{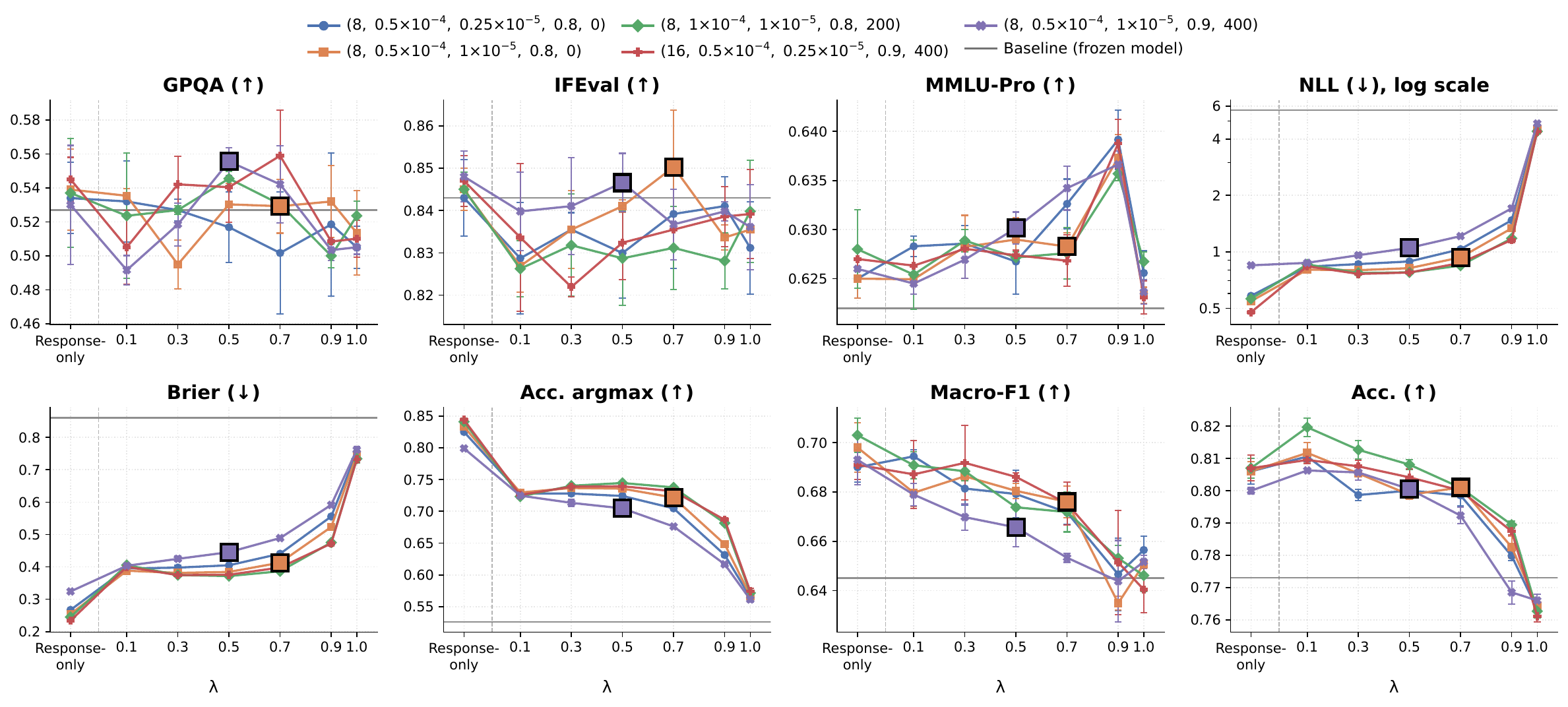}
  \caption{Single-adapter reasoning (Stage~2) across $\lambda$ values, initialized from the top response-only runs of Table~\ref{tab:response_only_results}.
  Each curve is one hyperparameter configuration $(\alpha,\ \text{LR},\ \text{min.\ LR},\ \beta_1,\ \text{warm-up})$;
  the leftmost point (``Response-only'') is the corresponding Stage~1 run. Points are means over three seeds and error bars show one standard deviation.
  Gray horizontal lines mark the frozen baseline model. Bold squares ($\blacksquare$) mark the two selected top runs,
  which improve over the baseline on every retention and in-domain metric.}
  \label{fig:lambda_ablation}
\end{figure}

Only two runs in this stage (bold squares, $\blacksquare$, in Figure~\ref{fig:lambda_ablation}) outperform the frozen model on all tasks. They remain behind the response-only runs they start from on macro-F1, argmax accuracy, NLL, and Brier.
Although the loss in Eq.~\eqref{eq:region_loss} is equivalent to the response-only objective at $\lambda=1.0$,
the $\lambda=1.0$ runs perform far worse than response-only training.
The reason is that the training inputs differ: in the reasoning setting, the model is trained on sequences that include the reasoning text,
which changes the context in which the label is learned.
We present the complete results in Appendix~\ref{app:single_adapter_results}.

\subsection{Dual-adapter Composite LoRA Results}\label{sec:exp_dual_adapter}

Dual-adapter training adds nothing beyond the single adapter. Against the response-only runs, calibration and argmax accuracy stay behind, as they do for the single adapter. Against the single adapter, retention behaves the same way, and the split of rank between the trace adapter and the label adapter does not separate the runs.
For each top run from Stage~2, we train dual-adapter models with the selected \(\lambda\) as well as two adjacent \(\lambda\) values. The number of trainable parameters is kept fixed and the total rank is \(r = 32\), as in the previous stages. Following Section~\ref{sec:method_dual_adapter}, we evaluate the splits
\((r_{\mathrm{trace}}, r_{\mathrm{label}}) = (16, 16)\), \((24, 8)\), and \((8, 24)\)
for each run, with \(\alpha/r\) held fixed per adapter.

\begin{figure}[!t]
  \centering
  \includegraphics[width=\linewidth]{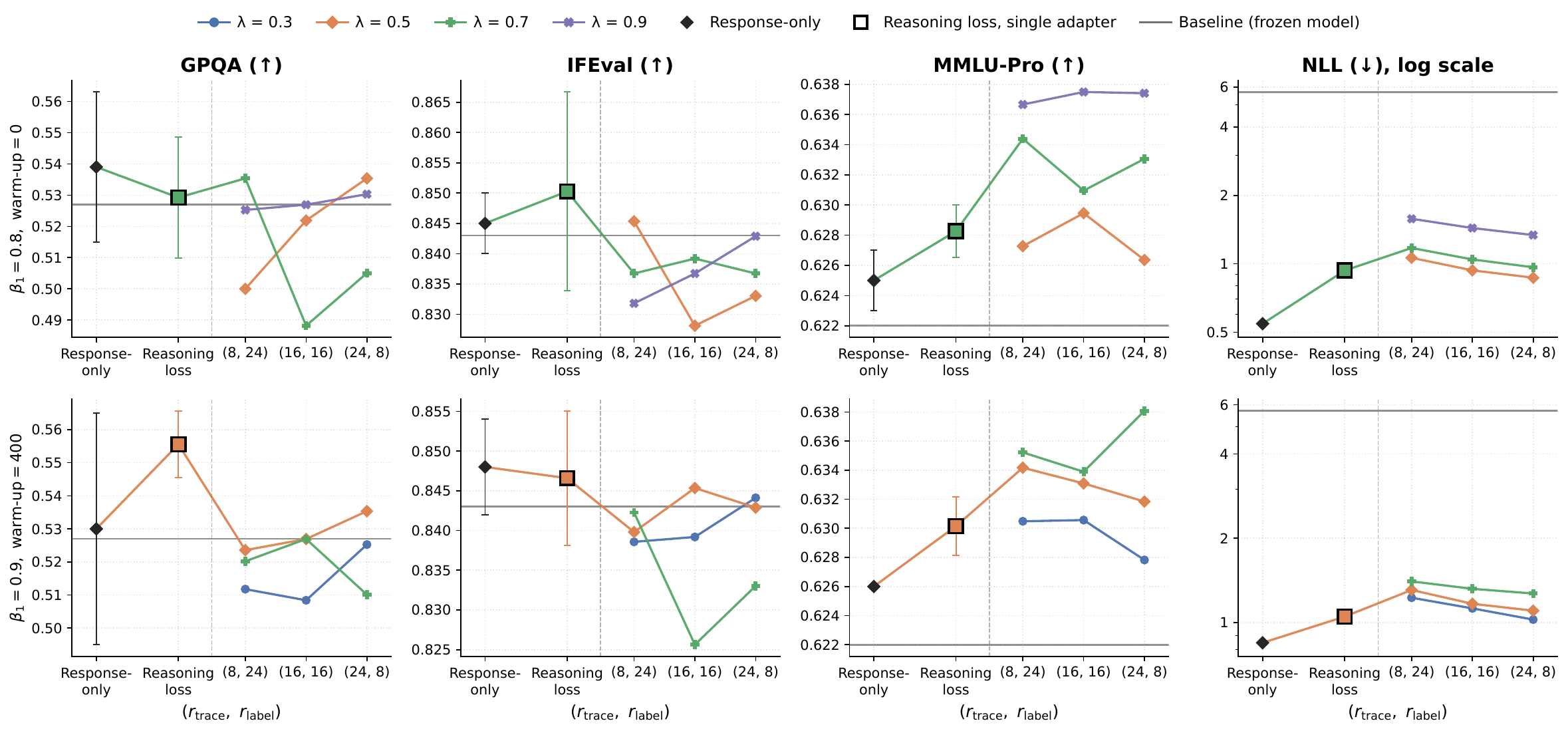}
  \caption{Dual-adapter composite stage (Stage~3). Each row corresponds to one top run from the
  single-adapter reasoning stage, trained with different optimizer settings, and is evaluated
  at the same \(\lambda\) as that run and at adjacent \(\lambda\) values. All runs share
  \(\alpha = 8\), a learning rate of \(5 \times 10^{-5}\), and a minimum learning rate of
  \(1 \times 10^{-5}\).}
  \label{fig:dual_adapter_ablation}
\end{figure}

Figure~\ref{fig:dual_adapter_ablation} shows this comparison on the retention benchmarks and on the in-domain metrics.
MMLU-Pro remains the retention benchmark that moves, as it does for the single adapter. IFEval stays near the frozen model and near the single-adapter and response-only runs. GPQA differences stay within the run-to-run variation of the frozen model.
NLL is lowest for response-only training, whose objective is the label itself and so matches the in-domain task directly. Brier, argmax accuracy, macro-F1, and greedy accuracy follow the same pattern. The full results are in Appendix~\ref{app:dual_adapter_results}.

\subsection{Summarized Traces Results}\label{sec:exp_summarized_traces}

In the fourth stage, we study the effect of training on summarized reasoning traces.
To construct length-controlled summaries, we compress each teacher-generated trace
with the teacher model itself (Qwen3-32B, greedy decoding) under a token-level length budget.
Given a trace of $T$ tokens and a target length ratio $K \in (0, 1]$,
we instruct the teacher to rewrite the trace as a condensed version of approximately $\lfloor K \cdot T \rceil$ tokens,
where $\lfloor \cdot \rceil$ denotes rounding to the nearest integer.
The summary must preserve the essential reasoning steps, the decisive evidence, and the final category choice,
while keeping the first-person deliberative style of the original trace (the exact prompt and implementation details are given in Appendix~\ref{app:compression_details}).
To prevent runaway generations without forcing mid-sentence truncation,
we cap the output at $1.1 \cdot \lfloor K \cdot T \rceil + 32$ tokens, i.e., $10\%$ above the target length plus a fixed margin of $32$ tokens. 
Table~\ref{tab:compression_ratios} shows the results of applying summarization on our dataset with different ratios.

\begin{table}[H]
\centering
\caption{Target vs.\ realized trace-compression ratios of the compressed
training sets, measured in tokens on 200K dataset. $\hat{K}$ is the per-trace ratio of
compressed to original trace length (mean $\pm$ std); the aggregate ratio is
the total compressed token count divided by the total original token count.
The first row gives the uncompressed source traces.}
\label{tab:compression_ratios}
\begin{tabular}{cccc}
\toprule
Target $K$ & $\hat{K}$ (mean $\pm$ std) & Aggregate token ratio & Trace tokens (mean $\pm$ std) \\
\midrule
Original & ---              & $1.000$ & $474 \pm 315$ \\
0.25 & $0.252 \pm 0.060$ & $0.233$ & $110 \pm 57$  \\
0.50 & $0.426 \pm 0.070$ & $0.414$ & $196 \pm 129$ \\
0.75 & $0.582 \pm 0.087$ & $0.576$ & $273 \pm 193$ \\
\bottomrule
\end{tabular}
\end{table}

\begin{figure}[H]
  \centering
  \includegraphics[width=\linewidth]{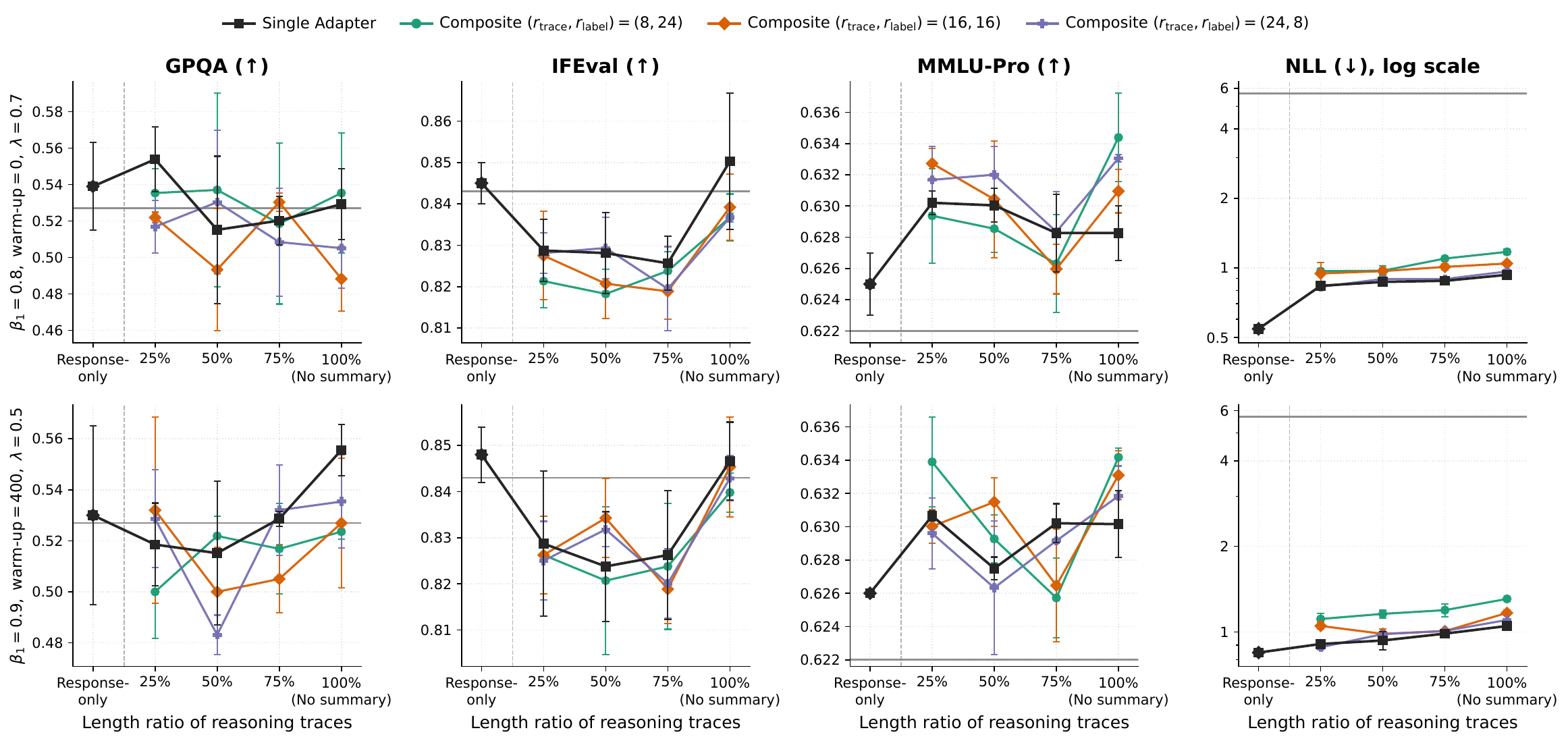}
  \caption{Effect of summarizing the reasoning traces (Stage~4) in the top runs of Stages~2 and~3.
  The $x$-axis is the target length of the summarized trace relative to the original ($100\%$ = no summary).
  The two runs are $\beta_1{=}0.8$ with no warm-up ($\lambda{=}0.7$) and $\beta_1{=}0.9$ with $400$ warm-up steps ($\lambda{=}0.5$), and
  use $\alpha{=}8$, LR $0.5{\times}10^{-4}$ and min.\ LR $1{\times}10^{-5}$.
  Points are means over three seeds, error bars show one standard deviation, and the gray line is the frozen baseline.}
  \label{fig:summary_ablation}
\end{figure}

Figure~\ref{fig:summary_ablation} shows training on compressed traces for the single adapter and the composite adapter. The horizontal axis is the target length. The 25\% target is reached; the 50\% and 75\% targets come out shorter than requested (Table~\ref{tab:compression_ratios}). The prompt is in Appendix~\ref{app:compression_details}.
Shorter summaries move NLL and Brier toward the response-only run, which stays ahead on both. Macro-F1 stays about the same across lengths. Greedy accuracy is lower at the 50\% and 75\% targets and close to the full trace at the 25\% target. IFEval sits about two points lower on the summaries than on the full trace, in both carried runs, and the seed intervals overlap. MMLU-Pro and GPQA stay within the seed variation. The composite adapter follows the single adapter, and no summary reaches the response-only run on NLL or Brier.
The rank split is visible mainly in NLL, where more rank on the label adapter tends to leave NLL worse. The detailed results are in Appendix~\ref{app:compression_results}.

\section{Conclusion}\label{sec:conclusion}

We presented a systematic study of reasoning supervision in LoRA fine-tuning for a financial classification task,
comparing response-only training with two ways of using teacher reasoning traces:
blending the trace into the loss of a single adapter and splitting the rank budget between a trace adapter and a label adapter.
We also studied the effect of summarizing the traces before training.
All regimes were compared under a staged protocol with matched hyperparameters and evaluated on both the in-domain task and retention benchmarks.

Response-only LoRA is the regime that improves in-domain performance while leaving general capabilities near the frozen model.
The gain is largest for calibration and for which class the model ranks first. Adding a full teacher trace does not extend this gain.
The in-domain metrics stay behind the response-only runs, MMLU-Pro can rise slightly,
and the other retention benchmarks stay within ordinary variation.
Splitting the rank across a trace adapter and a label adapter does not change this comparison.
The allocation only affects NLL, which worsens as more rank is given to the label adapter.
Summarizing the traces moves calibration back toward the response-only regime without reaching it,
lowers IFEval by a few points, and reduces accuracy and macro-F1 at intermediate summary lengths.
The composite adapter follows the single adapter at every summary length.

Several extensions remain open.
The same comparison can be repeated at other data scales and on other classification-like financial tasks, such as risk and compliance language or questions over filings.
Attaching a trace to only part of the training examples would test whether the MMLU-Pro lift can be kept at a smaller cost to calibration.
Decision models in the style of Jev~\citep{almeida2026jev} return one choice from a fixed list together with a probability.
They address the same problems and have not been compared with these LoRA regimes.
Finally, all of our experiments use AdamW, and the comparison could be repeated with other optimizers.

\let\origthebibliography\thebibliography
\renewcommand{\thebibliography}[1]{%
  \vspace{\baselineskip}\vspace{0pt}%
  \origthebibliography{#1}%
  \small
}
\bibliographystyle{unsrtnat}
\bibliography{references}

\clearpage
\appendix
\section{Appendix}\label{app:appendix}

\subsection{Hyperparameter Search for Response-only Stage}\label{app:hparam_search}

In this stage, we perform an extensive hyperparameter search. Table~\ref{tab:hparam_grid} summarizes
the fixed and searched hyperparameters used in this stage. The full grid amounts to 432 training runs for
response-only hyperparameter tuning.

\begin{table}[h]
\centering
\caption{Hyperparameters for response-only LoRA.
For Learning rate, we use linear warm-up followed by cosine decay to a floor (Min.\ learning rate). Each experiment
has 781 training steps.}
\label{tab:hparam_grid}
\begin{tabular}{lll}
\toprule
 & Hyperparameter & Values \\
\midrule
\multirow{6}{*}{Fixed}
  & Training examples in Dataset & $\{ 200\text{K}\}$ \\
  & Seeds                  & $\{617,\, 1234,\, 2468\}$ \\
  & LoRA rank ($r$)        & 32 \\
  & Adam $\beta_2$         & 0.999 \\
  & Global batch size      & 256 \\
  & Micro batch size (MBS) & 2 \\
\midrule
\multirow{5}{*}{Searched}
  & LoRA $\alpha$          & $\{8,\, 16,\, 32\}$ \\
  & Learning rate          & $\{5\times10^{-5},\, 1\times10^{-4},\, 2\times10^{-4},\, 5\times10^{-4}\}$ \\
  & Min.\ learning rate    & $\{2.5\times10^{-6},\, 1\times10^{-5}\}$ \\
  & Adam $\beta_1$         & $\{0.8,\, 0.9\}$ \\
  & Warm-up steps ($w$)    & $\{0,\, 200,\, 400\}$ \\
\bottomrule
\end{tabular}
\end{table}

\subsection{Detailed Results for Response-only Stage}\label{app:response_only_results}

Table~\ref{tab:response_only_results} lists the top runs defined in Section~\ref{sec:exp_response_only}.

\begin{table}[h]
  \centering
  \caption{Top runs of the response-only stage (Stage~1). Each configuration is trained with three random seeds and we report the mean $\pm$ standard deviation.
  Hyperparameters follow Table~\ref{tab:hparam_grid}.}
  \label{tab:response_only_results}
  \scriptsize
  \setlength{\tabcolsep}{3pt}
  \resizebox{\textwidth}{!}{%
  \begin{tabular}{ccccc|ccc ccccc}
    \toprule
    \multicolumn{5}{c|}{Hyperparameters} & \multicolumn{3}{c}{Retention} & \multicolumn{5}{c}{In-domain task} \\
    \cmidrule(lr){1-5} \cmidrule(lr){6-8} \cmidrule(lr){9-13}
    $\alpha$ & LR ($\times10^{-4}$) & Min.\ LR ($\times10^{-5}$) & $\beta_1$ & Warm-up & GPQA ($\uparrow$) & IFEval ($\uparrow$) & MMLU-Pro ($\uparrow$) & Acc.\ ($\uparrow$) & Macro-F1 ($\uparrow$) & Acc.\ argmax ($\uparrow$) & NLL ($\downarrow$) & Brier ($\downarrow$) \\
    \midrule
    \multicolumn{5}{c|}{\emph{Baseline (Frozen Model)}} & $0.527 \pm 0.044$ & $0.843 \pm 0.011$ & $0.622 \pm 0.000$ & $0.773 \pm 0.002$ & $0.645 \pm 0.012$ & $0.526 \pm 0.000$ & $5.701 \pm 0.002$ & $0.860 \pm 0.000$ \\
    \midrule
    8 & 0.5 & 0.25 & 0.8 & 0 & $0.534 \pm 0.021$ & $0.843 \pm 0.009$ & $0.625 \pm 0.002$ & $0.806 \pm 0.004$ & $0.690 \pm 0.006$ & $0.825 \pm 0.002$ & $0.582 \pm 0.010$ & $0.267 \pm 0.002$ \\
    8 & 0.5 & 1 & 0.8 & 0 & $0.539 \pm 0.024$ & $0.845 \pm 0.005$ & $0.625 \pm 0.002$ & $0.806 \pm 0.003$ & $0.698 \pm 0.010$ & $0.833 \pm 0.002$ & $0.545 \pm 0.011$ & $0.254 \pm 0.002$ \\
    8 & 1 & 1 & 0.8 & 200 & $0.537 \pm 0.032$ & $0.845 \pm 0.004$ & $0.628 \pm 0.004$ & $0.807 \pm 0.003$ & $0.703 \pm 0.007$ & $0.841 \pm 0.001$ & $0.562 \pm 0.010$ & $0.245 \pm 0.004$ \\
    16 & 0.5 & 0.25 & 0.9 & 400 & $0.545 \pm 0.013$ & $0.847 \pm 0.006$ & $0.627 \pm 0.001$ & $0.807 \pm 0.004$ & $0.691 \pm 0.006$ & $0.844 \pm 0.003$ & $0.476 \pm 0.011$ & $0.234 \pm 0.004$ \\
    8 & 0.5 & 1 & 0.9 & 400 & $0.530 \pm 0.035$ & $0.848 \pm 0.006$ & $0.626 \pm 0.000$ & $0.800 \pm 0.001$ & $0.693 \pm 0.010$ & $0.799 \pm 0.003$ & $0.847 \pm 0.013$ & $0.324 \pm 0.005$ \\
    \bottomrule
  \end{tabular}}
\end{table}

\pagebreak

\subsection{Detailed Results for Single-adapter Reasoning Stage}\label{app:single_adapter_results}

Table~\ref{tab:single_adapter_results} presents the detailed runs defined in Section~\ref{sec:exp_single_adapter}.

\begin{table}[h]
  \centering
  \caption{Single-adapter reasoning stage (Stage~2; 200K dataset, rank $r$=32) for each of the top response-only configurations of Table~\ref{tab:response_only_results} and each $\lambda$. Each configuration is trained with three random seeds and we report the mean $\pm$ standard deviation; the \emph{Response-only} row of each block repeats the corresponding run of Table~\ref{tab:response_only_results}. Bold rows ($\blacksquare$) are the two top runs that outperform the baseline on all tasks.}
  \label{tab:single_adapter_results}
  \scriptsize
  \setlength{\tabcolsep}{3pt}
  \resizebox{\textwidth}{!}{%
  \begin{tabular}{c|ccc|ccccc}
    \toprule
    & \multicolumn{3}{c|}{Retention} & \multicolumn{5}{c}{In-domain task} \\
    \cmidrule(lr){2-4} \cmidrule(lr){5-9}
    $\lambda$ & GPQA ($\uparrow$) & IFEval ($\uparrow$) & MMLU-Pro ($\uparrow$) & Acc.\ ($\uparrow$) & Macro-F1 ($\uparrow$) & Acc.\ argmax ($\uparrow$) & NLL ($\downarrow$) & Brier ($\downarrow$) \\
    \midrule
    \emph{Baseline (Frozen Model)} & $0.527 \pm 0.044$ & $0.843 \pm 0.011$ & $0.622 \pm 0.000$ & $0.773 \pm 0.002$ & $0.645 \pm 0.012$ & $0.526 \pm 0.000$ & $5.701 \pm 0.002$ & $0.860 \pm 0.000$ \\
    \midrule
    \multicolumn{9}{l}{\emph{$\alpha$=8, LR=0.5$\times10^{-4}$, Min.\ LR=0.25$\times10^{-5}$, $\beta_1$=0.8, Warm-up=0}} \\
    Response-only & $0.534 \pm 0.021$ & $0.843 \pm 0.009$ & $0.625 \pm 0.002$ & $0.806 \pm 0.004$ & $0.690 \pm 0.006$ & $0.825 \pm 0.002$ & $0.582 \pm 0.010$ & $0.267 \pm 0.002$ \\
    0.1 & $0.532 \pm 0.029$ & $0.829 \pm 0.016$ & $0.628 \pm 0.001$ & $0.811 \pm 0.002$ & $0.694 \pm 0.003$ & $0.728 \pm 0.002$ & $0.834 \pm 0.005$ & $0.395 \pm 0.003$ \\
    0.3 & $0.527 \pm 0.008$ & $0.835 \pm 0.005$ & $0.629 \pm 0.002$ & $0.799 \pm 0.002$ & $0.681 \pm 0.008$ & $0.728 \pm 0.004$ & $0.860 \pm 0.028$ & $0.398 \pm 0.008$ \\
    0.5 & $0.517 \pm 0.025$ & $0.830 \pm 0.013$ & $0.627 \pm 0.004$ & $0.800 \pm 0.001$ & $0.679 \pm 0.012$ & $0.724 \pm 0.003$ & $0.889 \pm 0.001$ & $0.405 \pm 0.002$ \\
    0.7 & $0.502 \pm 0.044$ & $0.839 \pm 0.016$ & $0.633 \pm 0.003$ & $0.799 \pm 0.004$ & $0.672 \pm 0.006$ & $0.705 \pm 0.003$ & $1.033 \pm 0.021$ & $0.441 \pm 0.005$ \\
    0.9 & $0.519 \pm 0.052$ & $0.841 \pm 0.008$ & $0.639 \pm 0.004$ & $0.780 \pm 0.002$ & $0.647 \pm 0.018$ & $0.631 \pm 0.003$ & $1.474 \pm 0.018$ & $0.556 \pm 0.005$ \\
    1.0 & $0.505 \pm 0.015$ & $0.831 \pm 0.013$ & $0.626 \pm 0.003$ & $0.763 \pm 0.003$ & $0.656 \pm 0.007$ & $0.565 \pm 0.006$ & $4.653 \pm 0.103$ & $0.754 \pm 0.014$ \\
    \midrule
    \multicolumn{9}{l}{\emph{$\alpha$=8, LR=0.5$\times10^{-4}$, Min.\ LR=1$\times10^{-5}$, $\beta_1$=0.8, Warm-up=0}} \\
    Response-only & $0.539 \pm 0.024$ & $0.845 \pm 0.005$ & $0.625 \pm 0.002$ & $0.806 \pm 0.003$ & $0.698 \pm 0.010$ & $0.833 \pm 0.002$ & $0.545 \pm 0.011$ & $0.254 \pm 0.002$ \\
    0.1 & $0.535 \pm 0.005$ & $0.827 \pm 0.007$ & $0.625 \pm 0.001$ & $0.812 \pm 0.004$ & $0.680 \pm 0.007$ & $0.730 \pm 0.005$ & $0.803 \pm 0.024$ & $0.388 \pm 0.007$ \\
    0.3 & $0.495 \pm 0.017$ & $0.835 \pm 0.011$ & $0.628 \pm 0.004$ & $0.805 \pm 0.001$ & $0.686 \pm 0.002$ & $0.737 \pm 0.003$ & $0.798 \pm 0.014$ & $0.381 \pm 0.003$ \\
    0.5 & $0.530 \pm 0.015$ & $0.841 \pm 0.002$ & $0.629 \pm 0.003$ & $0.799 \pm 0.002$ & $0.680 \pm 0.004$ & $0.736 \pm 0.001$ & $0.818 \pm 0.006$ & $0.384 \pm 0.002$ \\
    $\blacksquare$ \textbf{0.7} & $\mathbf{0.529 \pm 0.019}$ & $\mathbf{0.850 \pm 0.016}$ & $\mathbf{0.628 \pm 0.002}$ & $\mathbf{0.801 \pm 0.002}$ & $\mathbf{0.676 \pm 0.008}$ & $\mathbf{0.722 \pm 0.002}$ & $\mathbf{0.934 \pm 0.016}$ & $\mathbf{0.413 \pm 0.004}$ \\
    0.9 & $0.532 \pm 0.026$ & $0.834 \pm 0.004$ & $0.637 \pm 0.003$ & $0.783 \pm 0.004$ & $0.635 \pm 0.004$ & $0.648 \pm 0.002$ & $1.335 \pm 0.016$ & $0.523 \pm 0.002$ \\
    1.0 & $0.513 \pm 0.030$ & $0.835 \pm 0.005$ & $0.623 \pm 0.001$ & $0.764 \pm 0.001$ & $0.651 \pm 0.002$ & $0.567 \pm 0.004$ & $4.551 \pm 0.079$ & $0.746 \pm 0.010$ \\
    \midrule
    \multicolumn{9}{l}{\emph{$\alpha$=8, LR=1$\times10^{-4}$, Min.\ LR=1$\times10^{-5}$, $\beta_1$=0.8, Warm-up=200}} \\
    Response-only & $0.537 \pm 0.032$ & $0.845 \pm 0.004$ & $0.628 \pm 0.004$ & $0.807 \pm 0.003$ & $0.703 \pm 0.007$ & $0.841 \pm 0.001$ & $0.562 \pm 0.010$ & $0.245 \pm 0.004$ \\
    0.1 & $0.524 \pm 0.045$ & $0.826 \pm 0.008$ & $0.625 \pm 0.004$ & $0.820 \pm 0.004$ & $0.691 \pm 0.007$ & $0.723 \pm 0.003$ & $0.857 \pm 0.005$ & $0.405 \pm 0.004$ \\
    0.3 & $0.527 \pm 0.003$ & $0.832 \pm 0.015$ & $0.629 \pm 0.001$ & $0.813 \pm 0.003$ & $0.688 \pm 0.005$ & $0.740 \pm 0.002$ & $0.771 \pm 0.007$ & $0.375 \pm 0.004$ \\
    0.5 & $0.545 \pm 0.005$ & $0.829 \pm 0.014$ & $0.627 \pm 0.001$ & $0.808 \pm 0.002$ & $0.674 \pm 0.001$ & $0.745 \pm 0.002$ & $0.777 \pm 0.010$ & $0.372 \pm 0.002$ \\
    0.7 & $0.530 \pm 0.005$ & $0.831 \pm 0.012$ & $0.628 \pm 0.003$ & $0.801 \pm 0.002$ & $0.672 \pm 0.010$ & $0.738 \pm 0.001$ & $0.845 \pm 0.015$ & $0.386 \pm 0.002$ \\
    0.9 & $0.500 \pm 0.009$ & $0.828 \pm 0.008$ & $0.636 \pm 0.001$ & $0.789 \pm 0.001$ & $0.653 \pm 0.006$ & $0.682 \pm 0.004$ & $1.180 \pm 0.021$ & $0.475 \pm 0.005$ \\
    1.0 & $0.524 \pm 0.011$ & $0.840 \pm 0.015$ & $0.627 \pm 0.001$ & $0.763 \pm 0.003$ & $0.646 \pm 0.008$ & $0.571 \pm 0.009$ & $4.394 \pm 0.137$ & $0.734 \pm 0.017$ \\
    \midrule
    \multicolumn{9}{l}{\emph{$\alpha$=16, LR=0.5$\times10^{-4}$, Min.\ LR=0.25$\times10^{-5}$, $\beta_1$=0.9, Warm-up=400}} \\
    Response-only & $0.545 \pm 0.013$ & $0.847 \pm 0.006$ & $0.627 \pm 0.001$ & $0.807 \pm 0.004$ & $0.691 \pm 0.006$ & $0.844 \pm 0.003$ & $0.476 \pm 0.011$ & $0.234 \pm 0.004$ \\
    0.1 & $0.505 \pm 0.027$ & $0.834 \pm 0.021$ & $0.626 \pm 0.002$ & $0.810 \pm 0.001$ & $0.687 \pm 0.017$ & $0.726 \pm 0.006$ & $0.837 \pm 0.024$ & $0.400 \pm 0.008$ \\
    0.3 & $0.542 \pm 0.020$ & $0.822 \pm 0.003$ & $0.628 \pm 0.001$ & $0.808 \pm 0.002$ & $0.692 \pm 0.019$ & $0.739 \pm 0.004$ & $0.762 \pm 0.009$ & $0.374 \pm 0.004$ \\
    0.5 & $0.540 \pm 0.025$ & $0.832 \pm 0.011$ & $0.627 \pm 0.001$ & $0.804 \pm 0.003$ & $0.686 \pm 0.002$ & $0.739 \pm 0.001$ & $0.776 \pm 0.004$ & $0.376 \pm 0.001$ \\
    0.7 & $0.559 \pm 0.033$ & $0.835 \pm 0.005$ & $0.627 \pm 0.003$ & $0.800 \pm 0.001$ & $0.675 \pm 0.011$ & $0.732 \pm 0.002$ & $0.873 \pm 0.011$ & $0.398 \pm 0.004$ \\
    0.9 & $0.508 \pm 0.011$ & $0.839 \pm 0.009$ & $0.639 \pm 0.003$ & $0.787 \pm 0.001$ & $0.651 \pm 0.026$ & $0.687 \pm 0.004$ & $1.158 \pm 0.043$ & $0.472 \pm 0.009$ \\
    1.0 & $0.510 \pm 0.013$ & $0.839 \pm 0.013$ & $0.623 \pm 0.002$ & $0.761 \pm 0.002$ & $0.640 \pm 0.012$ & $0.574 \pm 0.006$ & $4.389 \pm 0.139$ & $0.732 \pm 0.012$ \\
    \midrule
    \multicolumn{9}{l}{\emph{$\alpha$=8, LR=0.5$\times10^{-4}$, Min.\ LR=1$\times10^{-5}$, $\beta_1$=0.9, Warm-up=400}} \\
    Response-only & $0.530 \pm 0.035$ & $0.848 \pm 0.006$ & $0.626 \pm 0.000$ & $0.800 \pm 0.001$ & $0.693 \pm 0.010$ & $0.799 \pm 0.003$ & $0.847 \pm 0.013$ & $0.324 \pm 0.005$ \\
    0.1 & $0.492 \pm 0.011$ & $0.840 \pm 0.011$ & $0.625 \pm 0.001$ & $0.806 \pm 0.001$ & $0.679 \pm 0.005$ & $0.725 \pm 0.000$ & $0.874 \pm 0.011$ & $0.403 \pm 0.003$ \\
    0.3 & $0.519 \pm 0.015$ & $0.841 \pm 0.014$ & $0.627 \pm 0.002$ & $0.806 \pm 0.003$ & $0.670 \pm 0.006$ & $0.713 \pm 0.007$ & $0.959 \pm 0.036$ & $0.425 \pm 0.010$ \\
    $\blacksquare$ \textbf{0.5} & $\mathbf{0.556 \pm 0.010}$ & $\mathbf{0.847 \pm 0.008}$ & $\mathbf{0.630 \pm 0.002}$ & $\mathbf{0.801 \pm 0.002}$ & $\mathbf{0.666 \pm 0.009}$ & $\mathbf{0.705 \pm 0.003}$ & $\mathbf{1.050 \pm 0.014}$ & $\mathbf{0.445 \pm 0.005}$ \\
    0.7 & $0.542 \pm 0.028$ & $0.837 \pm 0.010$ & $0.634 \pm 0.003$ & $0.792 \pm 0.003$ & $0.653 \pm 0.002$ & $0.676 \pm 0.002$ & $1.213 \pm 0.009$ & $0.489 \pm 0.004$ \\
    0.9 & $0.503 \pm 0.008$ & $0.840 \pm 0.003$ & $0.637 \pm 0.002$ & $0.768 \pm 0.004$ & $0.644 \pm 0.020$ & $0.617 \pm 0.001$ & $1.706 \pm 0.063$ & $0.591 \pm 0.009$ \\
    1.0 & $0.505 \pm 0.005$ & $0.836 \pm 0.012$ & $0.624 \pm 0.002$ & $0.766 \pm 0.002$ & $0.652 \pm 0.003$ & $0.561 \pm 0.005$ & $4.841 \pm 0.103$ & $0.762 \pm 0.012$ \\
    \bottomrule
  \end{tabular}}
\end{table}

\pagebreak

\subsection{Detailed Results for Dual-adapter Composite Stage}\label{app:dual_adapter_results}

Table~\ref{tab:dual_adapter_results} presents the detailed runs defined in Section~\ref{sec:exp_dual_adapter}.

\begin{table}[h]
  \centering
  \caption{Dual-adapter composite stage (Stage~3) for the two selected configurations,
  for adjacent $\lambda$ values and different rank splits
  $(r_{\text{trace}}, r_{\text{label}})$.
  Each run is trained with three random seeds and we report the mean $\pm$ standard deviation.
  For reference, each block starts with the response-only run of that configuration (Table~\ref{tab:response_only_results})
  and its selected single-adapter run with the reasoning loss (Table~\ref{tab:single_adapter_results}).
  Both configurations use $\alpha$=8, LR=0.5$\times10^{-4}$ and Min.\ LR=1$\times10^{-5}$.}
  \label{tab:dual_adapter_results}
  \scriptsize
  \setlength{\tabcolsep}{3pt}
  \resizebox{\textwidth}{!}{%
  \begin{tabular}{cc|ccc|ccccc}
    \toprule
    & & \multicolumn{3}{c|}{Retention} & \multicolumn{5}{c}{In-domain task} \\
    \cmidrule(lr){3-5} \cmidrule(lr){6-10}
    $\lambda$ & $(r_{\text{trace}}, r_{\text{label}})$ & GPQA ($\uparrow$) & IFEval ($\uparrow$) & MMLU-Pro ($\uparrow$) & Acc.\ ($\uparrow$) & Macro-F1 ($\uparrow$) & Acc.\ argmax ($\uparrow$) & NLL ($\downarrow$) & Brier ($\downarrow$) \\
    \midrule
    \multicolumn{2}{c|}{\emph{Baseline (Frozen Model)}} & $0.527 \pm 0.044$ & $0.843 \pm 0.011$ & $0.622 \pm 0.000$ & $0.773 \pm 0.002$ & $0.645 \pm 0.012$ & $0.526 \pm 0.000$ & $5.701 \pm 0.002$ & $0.860 \pm 0.000$ \\
    \midrule
    \multicolumn{10}{l}{\emph{$\beta_1$=0.8, Warm-up=0}} \\
    \multicolumn{2}{c|}{Response-only} & $0.539 \pm 0.024$ & $0.845 \pm 0.005$ & $0.625 \pm 0.002$ & $0.806 \pm 0.003$ & $0.698 \pm 0.010$ & $0.833 \pm 0.002$ & $0.545 \pm 0.011$ & $0.254 \pm 0.002$ \\
    \multicolumn{2}{c|}{Reasoning loss, single adapter ($\lambda$=0.7)} & $0.529 \pm 0.019$ & $0.850 \pm 0.016$ & $0.628 \pm 0.002$ & $0.801 \pm 0.002$ & $0.676 \pm 0.008$ & $0.722 \pm 0.002$ & $0.934 \pm 0.016$ & $0.413 \pm 0.004$ \\
    \addlinespace[2pt]
    0.5 & (8, 24) & $0.500 \pm 0.032$ & $0.845 \pm 0.006$ & $0.627 \pm 0.001$ & $0.797 \pm 0.004$ & $0.672 \pm 0.003$ & $0.696 \pm 0.003$ & $1.062 \pm 0.010$ & $0.451 \pm 0.004$ \\
    0.5 & (16, 16) & $0.522 \pm 0.030$ & $0.828 \pm 0.014$ & $0.629 \pm 0.000$ & $0.798 \pm 0.002$ & $0.675 \pm 0.015$ & $0.717 \pm 0.002$ & $0.936 \pm 0.017$ & $0.419 \pm 0.004$ \\
    0.5 & (24, 8) & $0.535 \pm 0.032$ & $0.833 \pm 0.012$ & $0.626 \pm 0.002$ & $0.795 \pm 0.001$ & $0.675 \pm 0.008$ & $0.721 \pm 0.006$ & $0.869 \pm 0.030$ & $0.403 \pm 0.010$ \\
    \addlinespace[2pt]
    0.7 & (8, 24) & $0.535 \pm 0.033$ & $0.837 \pm 0.006$ & $0.634 \pm 0.003$ & $0.797 \pm 0.002$ & $0.666 \pm 0.006$ & $0.676 \pm 0.006$ & $1.174 \pm 0.031$ & $0.480 \pm 0.008$ \\
    0.7 & (16, 16) & $0.488 \pm 0.018$ & $0.839 \pm 0.008$ & $0.631 \pm 0.001$ & $0.796 \pm 0.002$ & $0.656 \pm 0.005$ & $0.698 \pm 0.005$ & $1.045 \pm 0.020$ & $0.447 \pm 0.006$ \\
    0.7 & (24, 8) & $0.505 \pm 0.022$ & $0.837 \pm 0.001$ & $0.633 \pm 0.000$ & $0.797 \pm 0.002$ & $0.675 \pm 0.011$ & $0.712 \pm 0.004$ & $0.965 \pm 0.020$ & $0.426 \pm 0.006$ \\
    \addlinespace[2pt]
    0.9 & (8, 24) & $0.525 \pm 0.022$ & $0.832 \pm 0.010$ & $0.637 \pm 0.002$ & $0.771 \pm 0.004$ & $0.641 \pm 0.013$ & $0.634 \pm 0.006$ & $1.578 \pm 0.005$ & $0.561 \pm 0.005$ \\
    0.9 & (16, 16) & $0.527 \pm 0.013$ & $0.837 \pm 0.008$ & $0.637 \pm 0.001$ & $0.777 \pm 0.003$ & $0.640 \pm 0.006$ & $0.643 \pm 0.007$ & $1.437 \pm 0.028$ & $0.539 \pm 0.005$ \\
    0.9 & (24, 8) & $0.530 \pm 0.005$ & $0.843 \pm 0.017$ & $0.637 \pm 0.002$ & $0.776 \pm 0.003$ & $0.634 \pm 0.008$ & $0.649 \pm 0.005$ & $1.338 \pm 0.032$ & $0.523 \pm 0.006$ \\
    \midrule
    \multicolumn{10}{l}{\emph{$\beta_1$=0.9, Warm-up=400}} \\
    \multicolumn{2}{c|}{Response-only} & $0.530 \pm 0.035$ & $0.848 \pm 0.006$ & $0.626 \pm 0.000$ & $0.800 \pm 0.001$ & $0.693 \pm 0.010$ & $0.799 \pm 0.003$ & $0.847 \pm 0.013$ & $0.324 \pm 0.005$ \\
    \multicolumn{2}{c|}{Reasoning loss, single adapter ($\lambda$=0.5)} & $0.556 \pm 0.010$ & $0.847 \pm 0.008$ & $0.630 \pm 0.002$ & $0.801 \pm 0.002$ & $0.666 \pm 0.009$ & $0.705 \pm 0.003$ & $1.050 \pm 0.014$ & $0.445 \pm 0.005$ \\
    \addlinespace[2pt]
    0.3 & (8, 24) & $0.512 \pm 0.018$ & $0.839 \pm 0.010$ & $0.630 \pm 0.001$ & $0.799 \pm 0.002$ & $0.667 \pm 0.022$ & $0.675 \pm 0.002$ & $1.227 \pm 0.016$ & $0.491 \pm 0.003$ \\
    0.3 & (16, 16) & $0.508 \pm 0.003$ & $0.839 \pm 0.006$ & $0.631 \pm 0.002$ & $0.802 \pm 0.001$ & $0.661 \pm 0.004$ & $0.688 \pm 0.001$ & $1.124 \pm 0.007$ & $0.466 \pm 0.001$ \\
    0.3 & (24, 8) & $0.525 \pm 0.010$ & $0.844 \pm 0.007$ & $0.628 \pm 0.001$ & $0.801 \pm 0.005$ & $0.667 \pm 0.008$ & $0.701 \pm 0.002$ & $1.025 \pm 0.003$ & $0.443 \pm 0.002$ \\
    \addlinespace[2pt]
    0.5 & (8, 24) & $0.524 \pm 0.003$ & $0.840 \pm 0.004$ & $0.634 \pm 0.001$ & $0.796 \pm 0.001$ & $0.664 \pm 0.010$ & $0.665 \pm 0.001$ & $1.308 \pm 0.020$ & $0.507 \pm 0.002$ \\
    0.5 & (16, 16) & $0.527 \pm 0.025$ & $0.845 \pm 0.011$ & $0.633 \pm 0.001$ & $0.795 \pm 0.003$ & $0.652 \pm 0.019$ & $0.684 \pm 0.002$ & $1.167 \pm 0.025$ & $0.475 \pm 0.005$ \\
    0.5 & (24, 8) & $0.535 \pm 0.018$ & $0.843 \pm 0.005$ & $0.632 \pm 0.002$ & $0.798 \pm 0.001$ & $0.660 \pm 0.005$ & $0.694 \pm 0.004$ & $1.103 \pm 0.013$ & $0.460 \pm 0.005$ \\
    \addlinespace[2pt]
    0.7 & (8, 24) & $0.520 \pm 0.045$ & $0.842 \pm 0.003$ & $0.635 \pm 0.002$ & $0.781 \pm 0.002$ & $0.647 \pm 0.006$ & $0.643 \pm 0.003$ & $1.403 \pm 0.017$ & $0.536 \pm 0.003$ \\
    0.7 & (16, 16) & $0.527 \pm 0.030$ & $0.826 \pm 0.006$ & $0.634 \pm 0.003$ & $0.789 \pm 0.001$ & $0.651 \pm 0.002$ & $0.659 \pm 0.002$ & $1.320 \pm 0.008$ & $0.515 \pm 0.003$ \\
    0.7 & (24, 8) & $0.510 \pm 0.009$ & $0.833 \pm 0.002$ & $0.638 \pm 0.001$ & $0.790 \pm 0.004$ & $0.648 \pm 0.010$ & $0.666 \pm 0.003$ & $1.270 \pm 0.021$ & $0.502 \pm 0.005$ \\
    \bottomrule
  \end{tabular}}
\end{table}

\subsection{Details of Trace-Compression}\label{app:compression_details}

Each compression request uses the prompt below, sent through the teacher's chat template with thinking mode disabled and greedy decoding.
For a trace of $T$ tokens and a target length ratio $K$, \texttt{\{target\_tokens\}} $= \mathrm{round}(K \cdot T)$,
and \texttt{\{target\_words\}} and \texttt{\{ratio\_pct\}} give the same budget as an approximate word count and as a percentage, respectively.
\texttt{\{trace\}} is the original reasoning trace.
Generation is hard-capped at $1.1 \cdot \mathrm{round}(K \cdot T) + 32$ tokens, and outputs that reach the cap are trimmed back to the last complete sentence.
Table~\ref{tab:compression_ratios} reports the resulting trace lengths for each length ratio on the 200K dataset.

\begin{quote}
\ttfamily\small\raggedright
Below is a reasoning trace that explains how a financial news headline was
classified into an event category. Rewrite it as a condensed version that
preserves the essential reasoning steps and the conclusion.\\[4pt]
Requirements:\\
- The condensed version must be approximately \{target\_tokens\} tokens long
(roughly \{target\_words\} words; about \{ratio\_pct\}\% of the original
length). Stay within 5\% of that target length. Your output will be cut off
shortly past it, so finish your reasoning and state the category choice
before the limit.\\
- Keep the same first-person deliberative style as the original.\\
- Preserve the decisive evidence and the final category choice.\\
- Output ONLY the condensed reasoning text, with no preamble, headers, or
quotation marks.\\[4pt]
Original reasoning trace:\\
-{}-{}-\\
\{trace\}\\
-{}-{}-
\end{quote}

\pagebreak

\subsection{Detailed Results for Summarized Traces Stage}\label{app:compression_results}

Tables~\ref{tab:compression_single_adapter}, \ref{tab:compression_dual_adapter_a}, and~\ref{tab:compression_dual_adapter_b} present the detailed runs defined in Section~\ref{sec:exp_summarized_traces}.

\begin{table}[!h]
  \centering
  \caption{Single-adapter reasoning with summarized traces (Stage~4), for the two top runs.
  The length ratio is the target length of the summarized trace relative to the original one ($100\%$ = no summary).
  \emph{Response-only} is the run of the same configuration trained without traces (Table~\ref{tab:response_only_results}).
  Each run is trained with three random seeds and we report the mean $\pm$ standard deviation. Both runs use $\alpha$=8, LR=0.5$\times10^{-4}$ and Min.\ LR=1$\times10^{-5}$.}
  \label{tab:compression_single_adapter}
  \scriptsize
  \setlength{\tabcolsep}{3pt}
  \resizebox{\textwidth}{!}{%
  \begin{tabular}{c|ccc|ccccc}
    \toprule
    & \multicolumn{3}{c|}{Retention} & \multicolumn{5}{c}{In-domain task} \\
    \cmidrule(lr){2-4} \cmidrule(lr){5-9}
    Length ratio & GPQA ($\uparrow$) & IFEval ($\uparrow$) & MMLU-Pro ($\uparrow$) & Acc.\ ($\uparrow$) & Macro-F1 ($\uparrow$) & Acc.\ argmax ($\uparrow$) & NLL ($\downarrow$) & Brier ($\downarrow$) \\
    \midrule
    \emph{Baseline (Frozen Model)} & $0.527 \pm 0.044$ & $0.843 \pm 0.011$ & $0.622 \pm 0.000$ & $0.773 \pm 0.002$ & $0.645 \pm 0.012$ & $0.526 \pm 0.000$ & $5.701 \pm 0.002$ & $0.860 \pm 0.000$ \\
    \midrule
    \multicolumn{9}{l}{\emph{$\beta_1$=0.8, Warm-up=0, $\lambda$=0.7}} \\
    Response-only & $0.539 \pm 0.024$ & $0.845 \pm 0.005$ & $0.625 \pm 0.002$ & $0.806 \pm 0.003$ & $0.698 \pm 0.010$ & $0.833 \pm 0.002$ & $0.545 \pm 0.011$ & $0.254 \pm 0.002$ \\
    25\% & $0.554 \pm 0.018$ & $0.829 \pm 0.007$ & $0.630 \pm 0.001$ & $0.799 \pm 0.004$ & $0.681 \pm 0.008$ & $0.769 \pm 0.002$ & $0.837 \pm 0.024$ & $0.353 \pm 0.005$ \\
    50\% & $0.515 \pm 0.040$ & $0.828 \pm 0.010$ & $0.630 \pm 0.001$ & $0.766 \pm 0.008$ & $0.657 \pm 0.013$ & $0.751 \pm 0.003$ & $0.870 \pm 0.016$ & $0.375 \pm 0.002$ \\
    75\% & $0.520 \pm 0.013$ & $0.826 \pm 0.006$ & $0.628 \pm 0.002$ & $0.714 \pm 0.003$ & $0.652 \pm 0.013$ & $0.736 \pm 0.006$ & $0.880 \pm 0.033$ & $0.393 \pm 0.006$ \\
    100\% (No summary) & $0.529 \pm 0.019$ & $0.850 \pm 0.016$ & $0.628 \pm 0.002$ & $0.801 \pm 0.002$ & $0.676 \pm 0.008$ & $0.722 \pm 0.002$ & $0.934 \pm 0.016$ & $0.413 \pm 0.004$ \\
    \midrule
    \multicolumn{9}{l}{\emph{$\beta_1$=0.9, Warm-up=400, $\lambda$=0.5}} \\
    Response-only & $0.530 \pm 0.035$ & $0.848 \pm 0.006$ & $0.626 \pm 0.000$ & $0.800 \pm 0.001$ & $0.693 \pm 0.010$ & $0.799 \pm 0.003$ & $0.847 \pm 0.013$ & $0.324 \pm 0.005$ \\
    25\% & $0.519 \pm 0.016$ & $0.829 \pm 0.016$ & $0.631 \pm 0.000$ & $0.789 \pm 0.005$ & $0.673 \pm 0.004$ & $0.763 \pm 0.003$ & $0.908 \pm 0.015$ & $0.365 \pm 0.002$ \\
    50\% & $0.515 \pm 0.028$ & $0.824 \pm 0.012$ & $0.627 \pm 0.001$ & $0.761 \pm 0.008$ & $0.672 \pm 0.012$ & $0.731 \pm 0.001$ & $0.934 \pm 0.070$ & $0.400 \pm 0.009$ \\
    75\% & $0.529 \pm 0.003$ & $0.826 \pm 0.014$ & $0.630 \pm 0.001$ & $0.703 \pm 0.007$ & $0.655 \pm 0.017$ & $0.717 \pm 0.004$ & $0.987 \pm 0.029$ & $0.427 \pm 0.007$ \\
    100\% (No summary) & $0.556 \pm 0.010$ & $0.847 \pm 0.008$ & $0.630 \pm 0.002$ & $0.801 \pm 0.002$ & $0.666 \pm 0.009$ & $0.705 \pm 0.003$ & $1.050 \pm 0.014$ & $0.445 \pm 0.005$ \\
    \bottomrule
  \end{tabular}}
\end{table}

\begin{table}[p]
  \centering
  \caption{Dual-adapter composite LoRA with summarized traces (Stage~4) for the top run with $\beta_1$=0.8, warm-up=0 ($\alpha$=8, LR=0.5$\times10^{-4}$, Min.\ LR=1$\times10^{-5}$), for every length ratio, $\lambda$ and rank split $(r_{\text{trace}}, r_{\text{label}})$. Length ratio $100\%$ corresponds to the original traces. Each block starts with the single-adapter run with the reasoning loss ($\lambda$=0.7) trained on traces of the same length; \emph{Response-only} is the run of the same configuration trained without traces (Table~\ref{tab:response_only_results}). Each run is trained with three random seeds and we report the mean $\pm$ standard deviation.}
  \label{tab:compression_dual_adapter_a}
  \scriptsize
  \setlength{\tabcolsep}{3pt}
  \resizebox{\textwidth}{!}{%
  \begin{tabular}{cc|ccc|ccccc}
    \toprule
    & & \multicolumn{3}{c|}{Retention} & \multicolumn{5}{c}{In-domain task} \\
    \cmidrule(lr){3-5} \cmidrule(lr){6-10}
    $\lambda$ & $(r_{\text{trace}}, r_{\text{label}})$ & GPQA ($\uparrow$) & IFEval ($\uparrow$) & MMLU-Pro ($\uparrow$) & Acc.\ ($\uparrow$) & Macro-F1 ($\uparrow$) & Acc.\ argmax ($\uparrow$) & NLL ($\downarrow$) & Brier ($\downarrow$) \\
    \midrule
    \multicolumn{2}{c|}{\emph{Baseline (Frozen Model)}} & $0.527 \pm 0.044$ & $0.843 \pm 0.011$ & $0.622 \pm 0.000$ & $0.773 \pm 0.002$ & $0.645 \pm 0.012$ & $0.526 \pm 0.000$ & $5.701 \pm 0.002$ & $0.860 \pm 0.000$ \\
    \multicolumn{2}{c|}{Response-only} & $0.539 \pm 0.024$ & $0.845 \pm 0.005$ & $0.625 \pm 0.002$ & $0.806 \pm 0.003$ & $0.698 \pm 0.010$ & $0.833 \pm 0.002$ & $0.545 \pm 0.011$ & $0.254 \pm 0.002$ \\
    \midrule
    \multicolumn{10}{l}{\emph{Length ratio 25\%}} \\
    \multicolumn{2}{c|}{Single adapter ($\lambda$=0.7)} & $0.554 \pm 0.018$ & $0.829 \pm 0.007$ & $0.630 \pm 0.001$ & $0.799 \pm 0.004$ & $0.681 \pm 0.008$ & $0.769 \pm 0.002$ & $0.837 \pm 0.024$ & $0.353 \pm 0.005$ \\
    \addlinespace[2pt]
    0.5 & (8, 24) & $0.507 \pm 0.031$ & $0.832 \pm 0.003$ & $0.631 \pm 0.002$ & $0.794 \pm 0.004$ & $0.666 \pm 0.020$ & $0.750 \pm 0.008$ & $0.932 \pm 0.028$ & $0.379 \pm 0.009$ \\
    0.5 & (16, 16) & $0.520 \pm 0.038$ & $0.827 \pm 0.005$ & $0.631 \pm 0.002$ & $0.793 \pm 0.004$ & $0.683 \pm 0.008$ & $0.773 \pm 0.003$ & $0.760 \pm 0.006$ & $0.337 \pm 0.000$ \\
    0.5 & (24, 8) & $0.547 \pm 0.013$ & $0.822 \pm 0.013$ & $0.631 \pm 0.001$ & $0.800 \pm 0.004$ & $0.686 \pm 0.005$ & $0.781 \pm 0.004$ & $0.696 \pm 0.027$ & $0.321 \pm 0.005$ \\
    \addlinespace[2pt]
    0.7 & (8, 24) & $0.535 \pm 0.013$ & $0.821 \pm 0.006$ & $0.629 \pm 0.003$ & $0.783 \pm 0.004$ & $0.656 \pm 0.008$ & $0.727 \pm 0.007$ & $0.969 \pm 0.022$ & $0.411 \pm 0.008$ \\
    0.7 & (16, 16) & $0.522 \pm 0.003$ & $0.827 \pm 0.011$ & $0.633 \pm 0.001$ & $0.793 \pm 0.002$ & $0.683 \pm 0.005$ & $0.754 \pm 0.006$ & $0.946 \pm 0.111$ & $0.379 \pm 0.014$ \\
    0.7 & (24, 8) & $0.517 \pm 0.015$ & $0.828 \pm 0.005$ & $0.632 \pm 0.002$ & $0.794 \pm 0.012$ & $0.667 \pm 0.013$ & $0.759 \pm 0.004$ & $0.828 \pm 0.024$ & $0.359 \pm 0.005$ \\
    \addlinespace[2pt]
    0.9 & (8, 24) & $0.544 \pm 0.037$ & $0.835 \pm 0.008$ & $0.630 \pm 0.002$ & $0.779 \pm 0.001$ & $0.649 \pm 0.013$ & $0.663 \pm 0.003$ & $1.528 \pm 0.133$ & $0.544 \pm 0.015$ \\
    0.9 & (16, 16) & $0.520 \pm 0.010$ & $0.834 \pm 0.007$ & $0.630 \pm 0.002$ & $0.783 \pm 0.004$ & $0.652 \pm 0.014$ & $0.670 \pm 0.007$ & $1.372 \pm 0.109$ & $0.518 \pm 0.019$ \\
    0.9 & (24, 8) & $0.530 \pm 0.035$ & $0.833 \pm 0.004$ & $0.630 \pm 0.002$ & $0.775 \pm 0.006$ & $0.661 \pm 0.007$ & $0.693 \pm 0.003$ & $1.225 \pm 0.109$ & $0.478 \pm 0.016$ \\
    \midrule
    \multicolumn{10}{l}{\emph{Length ratio 50\%}} \\
    \multicolumn{2}{c|}{Single adapter ($\lambda$=0.7)} & $0.515 \pm 0.040$ & $0.828 \pm 0.010$ & $0.630 \pm 0.001$ & $0.766 \pm 0.008$ & $0.657 \pm 0.013$ & $0.751 \pm 0.003$ & $0.870 \pm 0.016$ & $0.375 \pm 0.002$ \\
    \addlinespace[2pt]
    0.5 & (8, 24) & $0.532 \pm 0.011$ & $0.835 \pm 0.006$ & $0.633 \pm 0.001$ & $0.783 \pm 0.006$ & $0.659 \pm 0.004$ & $0.729 \pm 0.002$ & $0.943 \pm 0.022$ & $0.407 \pm 0.002$ \\
    0.5 & (16, 16) & $0.522 \pm 0.015$ & $0.835 \pm 0.013$ & $0.629 \pm 0.001$ & $0.763 \pm 0.008$ & $0.656 \pm 0.007$ & $0.750 \pm 0.003$ & $0.808 \pm 0.007$ & $0.369 \pm 0.001$ \\
    0.5 & (24, 8) & $0.492 \pm 0.029$ & $0.834 \pm 0.008$ & $0.629 \pm 0.002$ & $0.754 \pm 0.004$ & $0.662 \pm 0.002$ & $0.762 \pm 0.005$ & $0.727 \pm 0.018$ & $0.345 \pm 0.008$ \\
    \addlinespace[2pt]
    0.7 & (8, 24) & $0.537 \pm 0.053$ & $0.818 \pm 0.006$ & $0.629 \pm 0.002$ & $0.773 \pm 0.003$ & $0.645 \pm 0.011$ & $0.717 \pm 0.003$ & $0.973 \pm 0.047$ & $0.423 \pm 0.007$ \\
    0.7 & (16, 16) & $0.493 \pm 0.034$ & $0.821 \pm 0.008$ & $0.630 \pm 0.004$ & $0.782 \pm 0.006$ & $0.654 \pm 0.020$ & $0.722 \pm 0.003$ & $0.969 \pm 0.034$ & $0.415 \pm 0.006$ \\
    0.7 & (24, 8) & $0.530 \pm 0.039$ & $0.829 \pm 0.007$ & $0.632 \pm 0.002$ & $0.767 \pm 0.007$ & $0.654 \pm 0.008$ & $0.744 \pm 0.001$ & $0.896 \pm 0.017$ & $0.388 \pm 0.003$ \\
    \addlinespace[2pt]
    0.9 & (8, 24) & $0.519 \pm 0.011$ & $0.824 \pm 0.012$ & $0.631 \pm 0.003$ & $0.765 \pm 0.005$ & $0.646 \pm 0.010$ & $0.666 \pm 0.001$ & $1.315 \pm 0.029$ & $0.522 \pm 0.005$ \\
    0.9 & (16, 16) & $0.517 \pm 0.024$ & $0.830 \pm 0.011$ & $0.634 \pm 0.002$ & $0.749 \pm 0.003$ & $0.645 \pm 0.006$ & $0.676 \pm 0.006$ & $1.239 \pm 0.064$ & $0.500 \pm 0.012$ \\
    0.9 & (24, 8) & $0.532 \pm 0.015$ & $0.820 \pm 0.002$ & $0.631 \pm 0.002$ & $0.749 \pm 0.002$ & $0.638 \pm 0.015$ & $0.681 \pm 0.004$ & $1.136 \pm 0.049$ & $0.481 \pm 0.010$ \\
    \midrule
    \multicolumn{10}{l}{\emph{Length ratio 75\%}} \\
    \multicolumn{2}{c|}{Single adapter ($\lambda$=0.7)} & $0.520 \pm 0.013$ & $0.826 \pm 0.006$ & $0.628 \pm 0.002$ & $0.714 \pm 0.003$ & $0.652 \pm 0.013$ & $0.736 \pm 0.006$ & $0.880 \pm 0.033$ & $0.393 \pm 0.006$ \\
    \addlinespace[2pt]
    0.5 & (8, 24) & $0.524 \pm 0.019$ & $0.821 \pm 0.016$ & $0.624 \pm 0.002$ & $0.727 \pm 0.005$ & $0.653 \pm 0.010$ & $0.714 \pm 0.001$ & $0.964 \pm 0.024$ & $0.425 \pm 0.003$ \\
    0.5 & (16, 16) & $0.534 \pm 0.015$ & $0.824 \pm 0.005$ & $0.625 \pm 0.002$ & $0.722 \pm 0.013$ & $0.627 \pm 0.019$ & $0.740 \pm 0.004$ & $0.842 \pm 0.018$ & $0.383 \pm 0.006$ \\
    0.5 & (24, 8) & $0.529 \pm 0.013$ & $0.826 \pm 0.014$ & $0.623 \pm 0.003$ & $0.711 \pm 0.006$ & $0.635 \pm 0.004$ & $0.747 \pm 0.001$ & $0.799 \pm 0.012$ & $0.373 \pm 0.002$ \\
    \addlinespace[2pt]
    0.7 & (8, 24) & $0.519 \pm 0.044$ & $0.824 \pm 0.005$ & $0.626 \pm 0.003$ & $0.745 \pm 0.001$ & $0.649 \pm 0.012$ & $0.697 \pm 0.003$ & $1.098 \pm 0.026$ & $0.462 \pm 0.004$ \\
    0.7 & (16, 16) & $0.530 \pm 0.005$ & $0.819 \pm 0.007$ & $0.626 \pm 0.002$ & $0.732 \pm 0.006$ & $0.641 \pm 0.012$ & $0.711 \pm 0.008$ & $1.011 \pm 0.031$ & $0.433 \pm 0.008$ \\
    0.7 & (24, 8) & $0.508 \pm 0.030$ & $0.819 \pm 0.010$ & $0.628 \pm 0.003$ & $0.725 \pm 0.011$ & $0.638 \pm 0.023$ & $0.726 \pm 0.004$ & $0.896 \pm 0.013$ & $0.404 \pm 0.001$ \\
    \addlinespace[2pt]
    0.9 & (8, 24) & $0.535 \pm 0.009$ & $0.835 \pm 0.008$ & $0.631 \pm 0.002$ & $0.738 \pm 0.006$ & $0.660 \pm 0.019$ & $0.655 \pm 0.003$ & $1.344 \pm 0.075$ & $0.535 \pm 0.010$ \\
    0.9 & (16, 16) & $0.495 \pm 0.027$ & $0.845 \pm 0.005$ & $0.632 \pm 0.003$ & $0.730 \pm 0.002$ & $0.648 \pm 0.006$ & $0.656 \pm 0.005$ & $1.422 \pm 0.100$ & $0.542 \pm 0.017$ \\
    0.9 & (24, 8) & $0.525 \pm 0.057$ & $0.829 \pm 0.005$ & $0.632 \pm 0.002$ & $0.727 \pm 0.001$ & $0.637 \pm 0.019$ & $0.661 \pm 0.003$ & $1.346 \pm 0.035$ & $0.528 \pm 0.006$ \\
    \midrule
    \multicolumn{10}{l}{\emph{Length ratio 100\% (no summary)}} \\
    \multicolumn{2}{c|}{Single adapter ($\lambda$=0.7)} & $0.529 \pm 0.019$ & $0.850 \pm 0.016$ & $0.628 \pm 0.002$ & $0.801 \pm 0.002$ & $0.676 \pm 0.008$ & $0.722 \pm 0.002$ & $0.934 \pm 0.016$ & $0.413 \pm 0.004$ \\
    \addlinespace[2pt]
    0.5 & (8, 24) & $0.500 \pm 0.032$ & $0.845 \pm 0.006$ & $0.627 \pm 0.001$ & $0.797 \pm 0.004$ & $0.672 \pm 0.003$ & $0.696 \pm 0.003$ & $1.062 \pm 0.010$ & $0.451 \pm 0.004$ \\
    0.5 & (16, 16) & $0.522 \pm 0.030$ & $0.828 \pm 0.014$ & $0.629 \pm 0.000$ & $0.798 \pm 0.002$ & $0.675 \pm 0.015$ & $0.717 \pm 0.002$ & $0.936 \pm 0.017$ & $0.419 \pm 0.004$ \\
    0.5 & (24, 8) & $0.535 \pm 0.032$ & $0.833 \pm 0.012$ & $0.626 \pm 0.002$ & $0.795 \pm 0.001$ & $0.675 \pm 0.008$ & $0.721 \pm 0.006$ & $0.869 \pm 0.030$ & $0.403 \pm 0.010$ \\
    \addlinespace[2pt]
    0.7 & (8, 24) & $0.535 \pm 0.033$ & $0.837 \pm 0.006$ & $0.634 \pm 0.003$ & $0.797 \pm 0.002$ & $0.666 \pm 0.006$ & $0.676 \pm 0.006$ & $1.174 \pm 0.031$ & $0.480 \pm 0.008$ \\
    0.7 & (16, 16) & $0.488 \pm 0.018$ & $0.839 \pm 0.008$ & $0.631 \pm 0.001$ & $0.796 \pm 0.002$ & $0.656 \pm 0.005$ & $0.698 \pm 0.005$ & $1.045 \pm 0.020$ & $0.447 \pm 0.006$ \\
    0.7 & (24, 8) & $0.505 \pm 0.022$ & $0.837 \pm 0.001$ & $0.633 \pm 0.000$ & $0.797 \pm 0.002$ & $0.675 \pm 0.011$ & $0.712 \pm 0.004$ & $0.965 \pm 0.020$ & $0.426 \pm 0.006$ \\
    \addlinespace[2pt]
    0.9 & (8, 24) & $0.525 \pm 0.022$ & $0.832 \pm 0.010$ & $0.637 \pm 0.002$ & $0.771 \pm 0.004$ & $0.641 \pm 0.013$ & $0.634 \pm 0.006$ & $1.578 \pm 0.005$ & $0.561 \pm 0.005$ \\
    0.9 & (16, 16) & $0.527 \pm 0.013$ & $0.837 \pm 0.008$ & $0.637 \pm 0.001$ & $0.777 \pm 0.003$ & $0.640 \pm 0.006$ & $0.643 \pm 0.007$ & $1.437 \pm 0.028$ & $0.539 \pm 0.005$ \\
    0.9 & (24, 8) & $0.530 \pm 0.005$ & $0.843 \pm 0.017$ & $0.637 \pm 0.002$ & $0.776 \pm 0.003$ & $0.634 \pm 0.008$ & $0.649 \pm 0.005$ & $1.338 \pm 0.032$ & $0.523 \pm 0.006$ \\
    \bottomrule
  \end{tabular}}
\end{table}

\begin{table}[p]
  \centering
  \caption{Dual-adapter composite LoRA with summarized traces (Stage~4) for the top run with $\beta_1$=0.9, warm-up=400 ($\alpha$=8, LR=0.5$\times10^{-4}$, Min.\ LR=1$\times10^{-5}$), for every length ratio, $\lambda$ and rank split $(r_{\text{trace}}, r_{\text{label}})$. Length ratio $100\%$ corresponds to the original traces. Each block starts with the single-adapter run with the reasoning loss ($\lambda$=0.5) trained on traces of the same length; \emph{Response-only} is the run of the same configuration trained without traces (Table~\ref{tab:response_only_results}). Each run is trained with three random seeds and we report the mean $\pm$ standard deviation.}
  \label{tab:compression_dual_adapter_b}
  \scriptsize
  \setlength{\tabcolsep}{3pt}
  \resizebox{\textwidth}{!}{%
  \begin{tabular}{cc|ccc|ccccc}
    \toprule
    & & \multicolumn{3}{c|}{Retention} & \multicolumn{5}{c}{In-domain task} \\
    \cmidrule(lr){3-5} \cmidrule(lr){6-10}
    $\lambda$ & $(r_{\text{trace}}, r_{\text{label}})$ & GPQA ($\uparrow$) & IFEval ($\uparrow$) & MMLU-Pro ($\uparrow$) & Acc.\ ($\uparrow$) & Macro-F1 ($\uparrow$) & Acc.\ argmax ($\uparrow$) & NLL ($\downarrow$) & Brier ($\downarrow$) \\
    \midrule
    \multicolumn{2}{c|}{\emph{Baseline (Frozen Model)}} & $0.527 \pm 0.044$ & $0.843 \pm 0.011$ & $0.622 \pm 0.000$ & $0.773 \pm 0.002$ & $0.645 \pm 0.012$ & $0.526 \pm 0.000$ & $5.701 \pm 0.002$ & $0.860 \pm 0.000$ \\
    \multicolumn{2}{c|}{Response-only} & $0.530 \pm 0.035$ & $0.848 \pm 0.006$ & $0.626 \pm 0.000$ & $0.800 \pm 0.001$ & $0.693 \pm 0.010$ & $0.799 \pm 0.003$ & $0.847 \pm 0.013$ & $0.324 \pm 0.005$ \\
    \midrule
    \multicolumn{10}{l}{\emph{Length ratio 25\%}} \\
    \multicolumn{2}{c|}{Single adapter ($\lambda$=0.5)} & $0.519 \pm 0.016$ & $0.829 \pm 0.016$ & $0.631 \pm 0.000$ & $0.789 \pm 0.005$ & $0.673 \pm 0.004$ & $0.763 \pm 0.003$ & $0.908 \pm 0.015$ & $0.365 \pm 0.002$ \\
    \addlinespace[2pt]
    0.3 & (8, 24) & $0.502 \pm 0.018$ & $0.834 \pm 0.008$ & $0.631 \pm 0.002$ & $0.792 \pm 0.004$ & $0.662 \pm 0.002$ & $0.707 \pm 0.001$ & $1.338 \pm 0.077$ & $0.451 \pm 0.007$ \\
    0.3 & (16, 16) & $0.517 \pm 0.032$ & $0.821 \pm 0.004$ & $0.630 \pm 0.001$ & $0.794 \pm 0.003$ & $0.679 \pm 0.008$ & $0.745 \pm 0.007$ & $0.975 \pm 0.088$ & $0.383 \pm 0.014$ \\
    0.3 & (24, 8) & $0.527 \pm 0.042$ & $0.827 \pm 0.004$ & $0.630 \pm 0.003$ & $0.794 \pm 0.003$ & $0.664 \pm 0.009$ & $0.751 \pm 0.008$ & $0.946 \pm 0.050$ & $0.374 \pm 0.012$ \\
    \addlinespace[2pt]
    0.5 & (8, 24) & $0.500 \pm 0.018$ & $0.826 \pm 0.008$ & $0.634 \pm 0.003$ & $0.789 \pm 0.004$ & $0.652 \pm 0.008$ & $0.717 \pm 0.003$ & $1.112 \pm 0.052$ & $0.430 \pm 0.006$ \\
    0.5 & (16, 16) & $0.532 \pm 0.037$ & $0.826 \pm 0.008$ & $0.630 \pm 0.001$ & $0.789 \pm 0.004$ & $0.657 \pm 0.009$ & $0.738 \pm 0.004$ & $1.051 \pm 0.028$ & $0.399 \pm 0.004$ \\
    0.5 & (24, 8) & $0.529 \pm 0.019$ & $0.825 \pm 0.009$ & $0.630 \pm 0.002$ & $0.787 \pm 0.009$ & $0.674 \pm 0.017$ & $0.759 \pm 0.001$ & $0.884 \pm 0.017$ & $0.365 \pm 0.003$ \\
    \addlinespace[2pt]
    0.7 & (8, 24) & $0.525 \pm 0.015$ & $0.834 \pm 0.009$ & $0.630 \pm 0.003$ & $0.777 \pm 0.004$ & $0.650 \pm 0.001$ & $0.692 \pm 0.005$ & $1.218 \pm 0.091$ & $0.472 \pm 0.013$ \\
    0.7 & (16, 16) & $0.519 \pm 0.013$ & $0.829 \pm 0.006$ & $0.629 \pm 0.001$ & $0.790 \pm 0.001$ & $0.663 \pm 0.009$ & $0.716 \pm 0.006$ & $1.033 \pm 0.039$ & $0.427 \pm 0.008$ \\
    0.7 & (24, 8) & $0.539 \pm 0.028$ & $0.826 \pm 0.006$ & $0.631 \pm 0.001$ & $0.784 \pm 0.002$ & $0.659 \pm 0.010$ & $0.722 \pm 0.004$ & $1.050 \pm 0.043$ & $0.420 \pm 0.008$ \\
    \midrule
    \multicolumn{10}{l}{\emph{Length ratio 50\%}} \\
    \multicolumn{2}{c|}{Single adapter ($\lambda$=0.5)} & $0.515 \pm 0.028$ & $0.824 \pm 0.012$ & $0.627 \pm 0.001$ & $0.761 \pm 0.008$ & $0.672 \pm 0.012$ & $0.731 \pm 0.001$ & $0.934 \pm 0.070$ & $0.400 \pm 0.009$ \\
    \addlinespace[2pt]
    0.3 & (8, 24) & $0.529 \pm 0.015$ & $0.834 \pm 0.014$ & $0.628 \pm 0.005$ & $0.774 \pm 0.006$ & $0.658 \pm 0.011$ & $0.705 \pm 0.004$ & $1.077 \pm 0.034$ & $0.438 \pm 0.003$ \\
    0.3 & (16, 16) & $0.545 \pm 0.030$ & $0.828 \pm 0.010$ & $0.628 \pm 0.001$ & $0.766 \pm 0.009$ & $0.683 \pm 0.014$ & $0.725 \pm 0.002$ & $1.030 \pm 0.016$ & $0.418 \pm 0.003$ \\
    0.3 & (24, 8) & $0.520 \pm 0.005$ & $0.828 \pm 0.011$ & $0.629 \pm 0.001$ & $0.766 \pm 0.006$ & $0.665 \pm 0.008$ & $0.731 \pm 0.005$ & $1.000 \pm 0.058$ & $0.407 \pm 0.009$ \\
    \addlinespace[2pt]
    0.5 & (8, 24) & $0.522 \pm 0.008$ & $0.821 \pm 0.016$ & $0.629 \pm 0.001$ & $0.778 \pm 0.003$ & $0.668 \pm 0.014$ & $0.694 \pm 0.000$ & $1.159 \pm 0.038$ & $0.459 \pm 0.005$ \\
    0.5 & (16, 16) & $0.500 \pm 0.017$ & $0.834 \pm 0.009$ & $0.631 \pm 0.001$ & $0.774 \pm 0.003$ & $0.662 \pm 0.011$ & $0.715 \pm 0.006$ & $0.985 \pm 0.038$ & $0.424 \pm 0.008$ \\
    0.5 & (24, 8) & $0.483 \pm 0.008$ & $0.832 \pm 0.004$ & $0.626 \pm 0.004$ & $0.759 \pm 0.005$ & $0.657 \pm 0.013$ & $0.722 \pm 0.003$ & $0.983 \pm 0.017$ & $0.417 \pm 0.002$ \\
    \addlinespace[2pt]
    0.7 & (8, 24) & $0.530 \pm 0.035$ & $0.826 \pm 0.007$ & $0.629 \pm 0.004$ & $0.758 \pm 0.003$ & $0.646 \pm 0.009$ & $0.673 \pm 0.006$ & $1.250 \pm 0.027$ & $0.500 \pm 0.004$ \\
    0.7 & (16, 16) & $0.513 \pm 0.029$ & $0.826 \pm 0.003$ & $0.630 \pm 0.002$ & $0.759 \pm 0.003$ & $0.648 \pm 0.011$ & $0.683 \pm 0.002$ & $1.206 \pm 0.064$ & $0.477 \pm 0.008$ \\
    0.7 & (24, 8) & $0.515 \pm 0.018$ & $0.823 \pm 0.009$ & $0.632 \pm 0.001$ & $0.762 \pm 0.005$ & $0.648 \pm 0.011$ & $0.695 \pm 0.001$ & $1.074 \pm 0.029$ & $0.455 \pm 0.006$ \\
    \midrule
    \multicolumn{10}{l}{\emph{Length ratio 75\%}} \\
    \multicolumn{2}{c|}{Single adapter ($\lambda$=0.5)} & $0.529 \pm 0.003$ & $0.826 \pm 0.014$ & $0.630 \pm 0.001$ & $0.703 \pm 0.007$ & $0.655 \pm 0.017$ & $0.717 \pm 0.004$ & $0.987 \pm 0.029$ & $0.427 \pm 0.007$ \\
    \addlinespace[2pt]
    0.3 & (8, 24) & $0.532 \pm 0.032$ & $0.829 \pm 0.008$ & $0.624 \pm 0.002$ & $0.734 \pm 0.006$ & $0.656 \pm 0.016$ & $0.697 \pm 0.005$ & $1.042 \pm 0.035$ & $0.448 \pm 0.007$ \\
    0.3 & (16, 16) & $0.513 \pm 0.012$ & $0.837 \pm 0.005$ & $0.628 \pm 0.005$ & $0.708 \pm 0.017$ & $0.646 \pm 0.011$ & $0.719 \pm 0.003$ & $0.925 \pm 0.026$ & $0.416 \pm 0.002$ \\
    0.3 & (24, 8) & $0.524 \pm 0.020$ & $0.829 \pm 0.003$ & $0.626 \pm 0.002$ & $0.713 \pm 0.015$ & $0.626 \pm 0.004$ & $0.733 \pm 0.001$ & $0.914 \pm 0.064$ & $0.402 \pm 0.008$ \\
    \addlinespace[2pt]
    0.5 & (8, 24) & $0.517 \pm 0.018$ & $0.824 \pm 0.014$ & $0.626 \pm 0.002$ & $0.729 \pm 0.006$ & $0.642 \pm 0.013$ & $0.678 \pm 0.003$ & $1.194 \pm 0.060$ & $0.484 \pm 0.007$ \\
    0.5 & (16, 16) & $0.505 \pm 0.013$ & $0.819 \pm 0.007$ & $0.626 \pm 0.003$ & $0.706 \pm 0.005$ & $0.650 \pm 0.006$ & $0.701 \pm 0.001$ & $1.008 \pm 0.020$ & $0.442 \pm 0.004$ \\
    0.5 & (24, 8) & $0.532 \pm 0.018$ & $0.820 \pm 0.007$ & $0.629 \pm 0.000$ & $0.704 \pm 0.002$ & $0.648 \pm 0.008$ & $0.708 \pm 0.003$ & $1.009 \pm 0.018$ & $0.437 \pm 0.002$ \\
    \addlinespace[2pt]
    0.7 & (8, 24) & $0.475 \pm 0.044$ & $0.826 \pm 0.013$ & $0.629 \pm 0.001$ & $0.721 \pm 0.008$ & $0.635 \pm 0.007$ & $0.659 \pm 0.002$ & $1.417 \pm 0.063$ & $0.534 \pm 0.009$ \\
    0.7 & (16, 16) & $0.513 \pm 0.028$ & $0.829 \pm 0.009$ & $0.632 \pm 0.001$ & $0.712 \pm 0.008$ & $0.652 \pm 0.009$ & $0.672 \pm 0.002$ & $1.169 \pm 0.026$ & $0.490 \pm 0.007$ \\
    0.7 & (24, 8) & $0.529 \pm 0.011$ & $0.827 \pm 0.003$ & $0.631 \pm 0.002$ & $0.706 \pm 0.008$ & $0.643 \pm 0.008$ & $0.676 \pm 0.005$ & $1.198 \pm 0.020$ & $0.489 \pm 0.004$ \\
    \midrule
    \multicolumn{10}{l}{\emph{Length ratio 100\% (no summary)}} \\
    \multicolumn{2}{c|}{Single adapter ($\lambda$=0.5)} & $0.556 \pm 0.010$ & $0.847 \pm 0.008$ & $0.630 \pm 0.002$ & $0.801 \pm 0.002$ & $0.666 \pm 0.009$ & $0.705 \pm 0.003$ & $1.050 \pm 0.014$ & $0.445 \pm 0.005$ \\
    \addlinespace[2pt]
    0.3 & (8, 24) & $0.512 \pm 0.018$ & $0.839 \pm 0.010$ & $0.630 \pm 0.001$ & $0.799 \pm 0.002$ & $0.667 \pm 0.022$ & $0.675 \pm 0.002$ & $1.227 \pm 0.016$ & $0.491 \pm 0.003$ \\
    0.3 & (16, 16) & $0.508 \pm 0.003$ & $0.839 \pm 0.006$ & $0.631 \pm 0.002$ & $0.802 \pm 0.001$ & $0.661 \pm 0.004$ & $0.688 \pm 0.001$ & $1.124 \pm 0.007$ & $0.466 \pm 0.001$ \\
    0.3 & (24, 8) & $0.525 \pm 0.010$ & $0.844 \pm 0.007$ & $0.628 \pm 0.001$ & $0.801 \pm 0.005$ & $0.667 \pm 0.008$ & $0.701 \pm 0.002$ & $1.025 \pm 0.003$ & $0.443 \pm 0.002$ \\
    \addlinespace[2pt]
    0.5 & (8, 24) & $0.524 \pm 0.003$ & $0.840 \pm 0.004$ & $0.634 \pm 0.001$ & $0.796 \pm 0.001$ & $0.664 \pm 0.010$ & $0.665 \pm 0.001$ & $1.308 \pm 0.020$ & $0.507 \pm 0.002$ \\
    0.5 & (16, 16) & $0.527 \pm 0.025$ & $0.845 \pm 0.011$ & $0.633 \pm 0.001$ & $0.795 \pm 0.003$ & $0.652 \pm 0.019$ & $0.684 \pm 0.002$ & $1.167 \pm 0.025$ & $0.475 \pm 0.005$ \\
    0.5 & (24, 8) & $0.535 \pm 0.018$ & $0.843 \pm 0.005$ & $0.632 \pm 0.002$ & $0.798 \pm 0.001$ & $0.660 \pm 0.005$ & $0.694 \pm 0.004$ & $1.103 \pm 0.013$ & $0.460 \pm 0.005$ \\
    \addlinespace[2pt]
    0.7 & (8, 24) & $0.520 \pm 0.045$ & $0.842 \pm 0.003$ & $0.635 \pm 0.002$ & $0.781 \pm 0.002$ & $0.647 \pm 0.006$ & $0.643 \pm 0.003$ & $1.403 \pm 0.017$ & $0.536 \pm 0.003$ \\
    0.7 & (16, 16) & $0.527 \pm 0.030$ & $0.826 \pm 0.006$ & $0.634 \pm 0.003$ & $0.789 \pm 0.001$ & $0.651 \pm 0.002$ & $0.659 \pm 0.002$ & $1.320 \pm 0.008$ & $0.515 \pm 0.003$ \\
    0.7 & (24, 8) & $0.510 \pm 0.009$ & $0.833 \pm 0.002$ & $0.638 \pm 0.001$ & $0.790 \pm 0.004$ & $0.648 \pm 0.010$ & $0.666 \pm 0.003$ & $1.270 \pm 0.021$ & $0.502 \pm 0.005$ \\
    \bottomrule
  \end{tabular}}
\end{table}

\end{document}